\documentclass[STIX1COL]{WileyNJD-v2}
\usepackage[square,numbers]{natbib}
\usepackage{tikz}
\usepackage{smartdiagram}
\usepackage{IEEEtrantools}
\usepackage{amsmath}
\usepackage{subfigure}
\usepackage{arydshln}
\newcommand{\dd}{\mathop{}\!\mathrm{d}}

\newtheorem{example}{Example}

\articletype{Article Type}%

\received{}
\revised{}
\accepted{}

\begin{document}

\title{Information Theoretic Evaluation of Privacy-Leakage, Interpretability, and Transferability for Trustworthy AI}

\author[1,2]{Mohit Kumar*}

\author[1]{Bernhard A. Moser}

\author[1]{Lukas Fischer}

\author[1]{Bernhard Freudenthaler}

\authormark{KUMAR \textsc{et al}}

\address[1]{\orgname{Software Competence Center Hagenberg}, \orgaddress{\state{Upper Austria}, \country{Austria}}}

\address[2]{\orgdiv{Faculty of Computer Science and Electrical Engineering}, \orgname{University of Rostock}, \orgaddress{\state{Mecklenburg-Vorpommern}, \country{Germany}}}

\corres{*Mohit Kumar, Softwarepark 32a, A-4232 Hagenberg, Austria. \email{mohit.kumar@uni-rostock.de}}


\abstract[Summary]{In order to develop machine learning and deep learning models that take into account the guidelines and principles of trustworthy AI, a novel information theoretic trustworthy AI framework is introduced. A unified approach to ``privacy-preserving interpretable and transferable learning'' is considered for studying and optimizing the tradeoffs between privacy, interpretability, and transferability aspects. A variational membership-mapping Bayesian model is used for the analytical approximations of the defined information theoretic measures for privacy-leakage, interpretability, and transferability. The approach consists of approximating the information theoretic measures via maximizing a lower-bound using variational optimization. The study presents a unified information theoretic approach to study different aspects of trustworthy AI in a rigorous analytical manner. The approach is demonstrated through numerous experiments on benchmark datasets and a real-world biomedical application concerned with the detection of mental stress on individuals using heart rate variability analysis.}


\maketitle

\footnotetext{\textbf{Abbreviations:} TAI, trustworthy AI}

\section{Introduction}
Trust in the development, deployment, and use of AI is essential to fully utilize the AI-potential in contributing to human well being and society. The recent advances in machine and deep learning have rejuvenated the field of AI with an enthusiasm that AI would become an integral part of human life. However, rapid proliferation of AI will give rise to several ethical, legal, and social issues. 
\subsection{Trustworthy AI}
In response to the ethical, legal, and social challenges accompanied by AI, guidelines and ethical principles have been established~\cite{ec2019ethics,Floridi2019,Floridi2019Unified,Floridi2018} to evaluate the responsible development of AI systems that are good for humanity and the environment. The guidelines have introduced the concept of \emph{trustworthy} AI (TAI) and the term TAI has quickly gained attention in research and practice. TAI is based on the idea that trust in AI will make AI realize its full potential in contributing to societies, economies, and sustainable development. As ``trust'' is a complex phenomenon being studied in diverse disciplines (i.e. psychology, sociology, economics, management, computer science, and information systems), the definition and realization of TAI remains challenging. While forming trust in technology, users express expectations about the technology's \emph{functionality}, \emph{helpfulness} and \emph{reliability}~\cite{10.1145/1985347.1985353}.
The authors in~\cite{Thiebes2020} state that ``\emph{AI is perceived as trustworthy by its users (e.g., consumers, organizations, society) when it is developed, deployed, and used in ways that not only ensure its compliance with all relevant laws and its robustness but especially its adherence to general ethical principles}''. 

Academicians, industries, and policymakers have developed in recent times for TAI several frameworks and guidelines including ``Asilomar AI Principles''~\cite{Asilomar2017}, ``Montreal Declaration of Responsible AI''~\cite{Montreal2017}, ``UK AI Code''~\cite{uk2017}, ``AI4People''~\cite{Floridi2018}, ``Ethics Guidelines for Trustworthy AI''~\cite{ec2019ethics}, ``OECD Principles on AI''~\cite{oecd2019}, ``Governance Principles for the New Generation Artificial Intelligence''~\cite{china2019}, and ``Guidance for Regulation of Artificial Intelligence Applications''~\cite{vought2020}. However, it was argued in~\cite{Hagendorff2020ER} that AI ethics lack a reinforcement mechanism and economic incentives could easily override commitment to ethical principles and values.

The five principles of ethical AI~\cite{Floridi2018} (i.e. \emph{beneficence}, \emph{non-maleficence}, \emph{autonomy}, \emph{justice}, and \emph{explicability}) have been adopted for TAI~\cite{Thiebes2020}. Beneficence refers to promoting well-being of humans, preserving dignity, and sustaining the planet. Non-maleficence refers to avoiding bringing harm to people and is especially concerned with the protection of people's privacy and security. Autonomy refers to the promotion of human autonomy, agency, and oversight including the restriction of AI Systems' autonomy, where necessary. Justice refers to using AI for correcting past wrongs, ensuring shared benefits through AI; and preventing the creation of new harms and inequities by AI. Explicability comprises an epistemological sense and an ethical sense. Explicability refers in epistemological sense to the explainable AI via creating interpretable AI models with high levels of performance and accuracy. In ethical sense, explicability refers to accountable AI. Despite the importance of outlined TAI principles, their major limitation, as identified in~\cite{Thiebes2020}, is concerning the fact that principles are highly general and provide little to no guidance for how they can be transferred into practice. To address this limitation, a data-driven research framework for TAI was outlined in~\cite{Thiebes2020}. 
 \subsection{Motivation of the Current Study}
The core issues related to machine and deep learning, that need to be addressed for fulfilling the five principles of trustworthy AI, are listed Table~\ref{table_TAI}. 
\begin{table}
\caption{Core issues of TAI principles and solution approach}
\label{table_TAI} \centering
\begin{tabular}{r||l|l}
\toprule
\bfseries TAI principle & \bfseries issue & \bfseries solution approach \\ \midrule
\multirow{2}{*}{Beneficence} & \textbf{I1}: $\begin{array}{l} \mbox{non-availability of large} \\ \mbox{high-quality training data} \end{array}$  & transfer learning \\ \cmidrule{2-3}
 &  \textbf{I2}: $\begin{array}{l} \mbox{models (intellectual properties)} \\ \mbox{are not widely available} \end{array}$ & federated learning  \\ \midrule
 \multirow{2}{*}{Non-maleficence}   &  \textbf{I3}: $\begin{array}{l}  \mbox{leakage of private information} \\ \mbox{embedded in training data} \end{array}$ &  $\begin{array}{l}  \mbox{privacy-preserving} \\ \mbox{data release mechanism} \end{array}$ \\ 
 \cmidrule{2-3}
 &  \textbf{I4}: $\begin{array}{l}  \mbox{leakage of private information} \\ \mbox{embedded in model parameters} \\ \mbox{and model outputs} \end{array}$ & $\begin{array}{l}  \mbox{privacy-preserving} \\ \mbox{machine and deep learning} \end{array}$  \\ \midrule
Autonomy &  \textbf{I5}: $\begin{array}{l}  \mbox{user's inability to quantify} \\ \mbox{model-uncertainties leads to} \\ \mbox{indecisiveness regarding the level} \\ \mbox{of autonomy given to AI system} \end{array}$ & $\begin{array}{l}  \mbox{analytical quantification of} \\ \mbox{model uncertainties} \end{array}$   \\ \midrule
Justice &  \textbf{I6}: $\begin{array}{l}  \mbox{bias of training data} \\ \mbox{towards certain groups of people} \\ \mbox{leads to discrimination} \end{array}$ & federated learning \\ 
\midrule
Explicability &  \textbf{I7}: $\begin{array}{l}  \mbox{user's inability to understand} \\ \mbox{model functionality leads} \\ \mbox{to mistrust and obstruction} \\ \mbox{in establishing accountability} \end{array}$ &  $\begin{array}{l}  \mbox{interpretable machine and} \\ \mbox{deep learning models} \end{array}$ \\
 \bottomrule
\end{tabular}
\end{table} 
The solution approaches to address the issues concerning TAI (as identified in Table~\ref{table_TAI}) do exist in the literature, however, a unified solution approach addressing all major issues doesn't exist. Thus, a novel trustworthy AI framework is proposed for addressing the core issues in a rigorous analytical manner.  
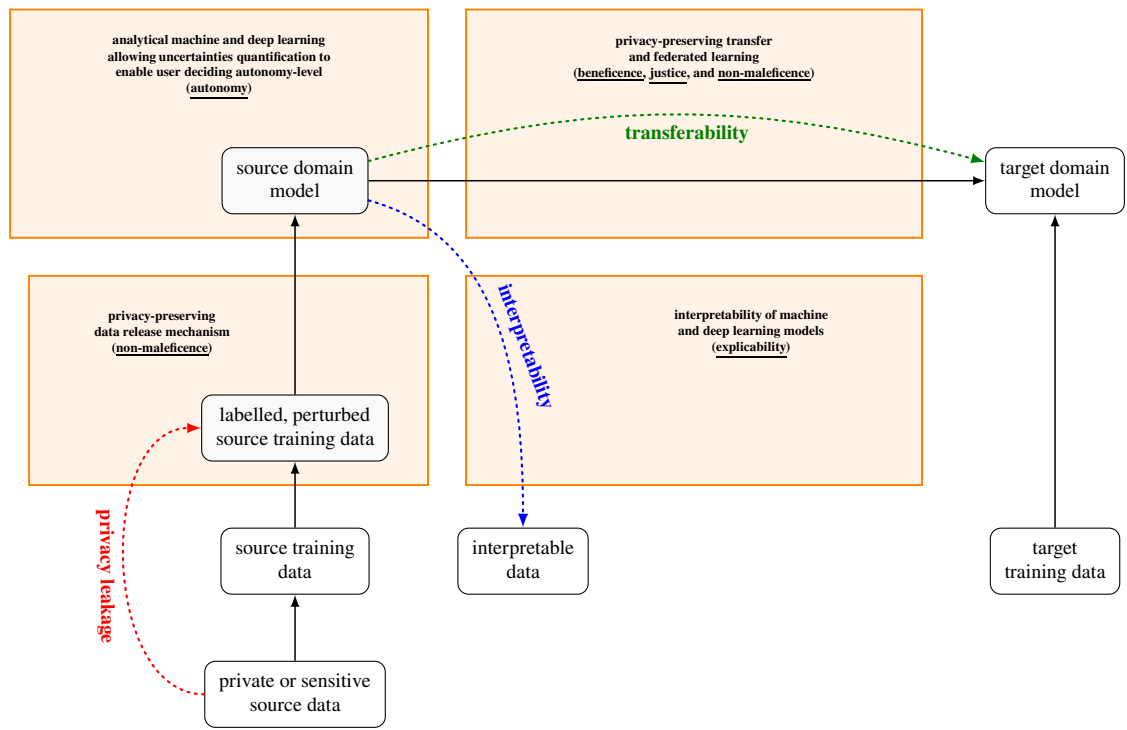
\begin{figure}
\definecolor{ao(english)}{rgb}{0.0, 0.5, 0.0}
\centering
\begin{tikzpicture}[scale=1]
\draw (0,0) node[rounded corners,draw](n1){\footnotesize $\begin{array}{c}\mbox{private or sensitive} \\ \mbox{source data} \end{array}$};
\path[fill=orange!10](-3.5,2.75)--(1.75,2.75)--(1.75,5.5)--(-3.5,5.5)--cycle;
  \draw[orange,line width = 0.25mm] (-3.5,2.75)--(1.75,2.75)--(1.75,5.5)--(-3.5,5.5)--cycle;
\draw (0,1.75) node[rounded corners,draw](n2){ \footnotesize $\begin{array}{c}\mbox{source training} \\ \mbox{data} \end{array}$};
\draw[-latex,line width=0.2mm] (n1) to [out=90,in=-90] (n2);  
\draw (0,3.5) node[rounded corners,draw, fill=gray!5](n3){\footnotesize $\begin{array}{c}\mbox{labelled, perturbed} \\ \mbox{source training data} \end{array}$};
\draw (-1.75,4.75) node[]{\bfseries \tiny $\begin{array}{c}\mbox{privacy-preserving} \\ \mbox{data release mechanism} \\ \mbox{(\underline{non-maleficence})} \end{array}$};
\draw[-latex,line width=0.2mm] (n2) to [out=90,in=-90] (n3);  
\path[fill=orange!10](-3.75,6)--(1.75,6)--(1.75,9)--(-3.75,9)--cycle;
  \draw[orange,line width = 0.25mm] (-3.75,6)--(1.75,6)--(1.75,9)--(-3.75,9)--cycle;
\draw (0,6.75) node[rounded corners,draw, fill=gray!5](n4){\footnotesize $\begin{array}{c}\mbox{source domain} \\ \mbox{model} \end{array}$};
\draw (-1,8.25) node[]{\bfseries \tiny $\begin{array}{c}\mbox{analytical machine and deep learning} \\ \mbox{allowing uncertainties quantification to} \\ \mbox{enable user deciding autonomy-level} \\ \mbox{(\underline{autonomy})}  \end{array}$};
\draw[-latex,line width=0.2mm] (n3) to [out=90,in=-90] (n4);  
\draw[-latex,dotted,thick,line cap=round,color=red] (n1) to [out=180,in=-180]  node[below,sloped,red] {\bfseries \footnotesize ~~~~~~privacy leakage} (n3); 
\draw (3,1.75) node[rounded corners,draw](n5){ \footnotesize $\begin{array}{c}\mbox{interpretable} \\ \mbox{data} \end{array}$}; 
\path[fill=orange!10](2.25,6)--(8.25,6)--(8.25,9)--(2.25,9)--cycle;  
\draw[orange,line width = 0.25mm] (2.25,6)--(8.25,6)--(8.25,9)--(2.25,9)--cycle;
\path[fill=orange!10](2.25,2.75)--(8.25,2.75)--(8.25,5.5)--(2.25,5.5)--cycle;  
\draw[orange,line width = 0.25mm](2.25,2.75)--(8.25,2.75)--(8.25,5.5)--(2.25,5.5)--cycle;     
\draw[-latex,dotted,thick,line cap=round,color=blue] (n4) to [out=-15,in=90]  node[above,sloped,blue] {\bfseries \footnotesize ~~~~~~~~~~~~interpretability} (n5); 
 \draw (10,1.75) node[rounded corners,draw](n6){\footnotesize $\begin{array}{c} \mbox{target} \\ \mbox{training data} \end{array}$};
 \draw (10,6.75) node[rounded corners,draw](n7){\footnotesize $\begin{array}{c}\mbox{target domain} \\ \mbox{model} \end{array}$};
 \draw (5.25,8.35) node[]{\bfseries \tiny $\begin{array}{c}\mbox{privacy-preserving transfer} \\ \mbox{and federated learning} \\ \mbox{(\underline{beneficence}, \underline{justice}, and \underline{non-maleficence})}  \end{array}$};
 \draw[-latex,line width=0.2mm] (n6) to [out=90,in=-90] (n7);  
  \draw[-latex,line width=0.2mm] (n4) to [out=0,in=180] (n7);  
  \draw[-latex,dotted,thick,line cap=round,color=ao(english)] (n4) to [out=15,in=165]  node[below,sloped,ao(english)] {\bfseries \footnotesize ~~~~transferability} (n7); 
\draw (6,4.75) node[]{\bfseries \tiny $\begin{array}{c}\mbox{interpretability of machine} \\ \mbox{and deep learning models} \\ \mbox{(\underline{explicability})} \end{array}$};  
\end{tikzpicture}
\caption{ITTAI framework facilitates a transfer of TAI principles (beneficence, non-maleficence, autonomy, justice, and explicability) into practice via providing an information theoretic unified approach to ``privacy-preserving interpretable and transferable learning'' for studying the privacy-interpretability-transferability tradeoffs.}\label{figure_TAI}
\end{figure}  
We introduce a novel framework, referred to as \emph{Information Theoretic Trustworthy Artificial Intelligence} (ITTAI), for the design and analysis of trustworthy AI systems. The ITTAI framework is based on the hypothesis that \emph{information theory enables taking into account the trustworthy AI principles of beneficence, non-maleficence, autonomy, justice, and explicability during the development of machine learning and deep learning based AI systems via providing a way to study and optimize the inherent tradeoffs between TAI principles}. The overall aim of ITTAI framework is to facilitate transfer of TAI principles into practice via fulfilling following aims:
\begin{description}
\item[Aim 1:] To develop an information theoretic approach to privacy enabling the quantification of privacy leakage in-terms of mutual information between sensitive private data and the released public data without the availability of a prior knowledge about data statistics (such as joint distributions of public and private variables).   
\item[Aim 2:] To develop an information theoretic criterion for evaluating the interpretability of a machine learning model in-terms of mutual information between non-interpretable model outputs/activations and corresponding interpretable parameters.  
\item[Aim 3:] To develop an information theoretic criterion for evaluating the transferability (of a machine learning model from source to target domain) in-terms of mutual information between source domain model outputs/activations and target domain model outputs/activations.    
\item[Aim 4:] To develop analytical approaches to machine and deep learning allowing quantification of model uncertainties.
\item[Aim 5:] To develop a unified approach to ``privacy-preserving interpretable and transferable learning'' for an analytical optimization of privacy-interpretability-transferability tradeoffs. 
\end{description}    
ITTAI framework (with its structure as in Fig.~\ref{figure_TAI}) addresses the
\begin{enumerate}
\item issues \textbf{I1} and \textbf{I2} of beneficence principle by means of transfer and federated learning;
\item issues \textbf{I3} and \textbf{I4} of non-maleficence principle by means of privacy-preserving data release mechanisms;
\item issue \textbf{I5} of autonomy principle by means of analytical machine and deep learning algorithms enabling the user to quantify model uncertainties and hence to decide the level of autonomy given to AI systems;  
\item issue \textbf{I6} of justice principle by means of federated learning;    
\item issue \textbf{I7} of explicability principle by means of interpretable machine and deep learning models.  
\end{enumerate}   
The most important feature of ITTAI is that the notions of privacy, interpretability, and transferability are quantified by means of information theoretic measures allowing the study and optimization of tradeoffs between TAI principles (such as tradeoff between privacy and transferability, or tradeoff between privacy and interpretability) in a practical manner.  
\subsection{Methodology}
\begin{figure}
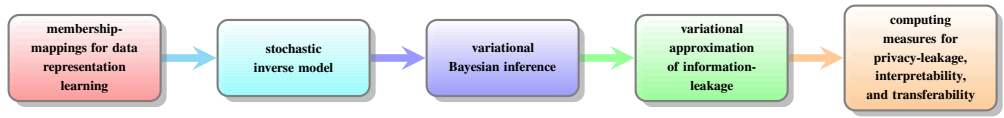

\centering
\smartdiagramset{back arrow disabled=true}
\smartdiagram[flow diagram:horizontal]{\textbf{membership-mappings for data representation learning},
 \textbf{stochastic inverse model}, \textbf{variational Bayesian inference}, \textbf{variational approximation of information-leakage}, {\textbf{computing measures for privacy-leakage, interpretability, and transferability}}}
  \caption{The proposed methodology to evaluate privacy-leakage, interpretability, and transferability in-terms of information-leakages.}\label{figure_methodology}
 \end{figure}

Fig.~\ref{figure_methodology} outlines the methodological workflow. For an information theoretic evaluation of privacy-leakage, interpretability, and transferability, the study provides a novel approach consisting of following three steps:
\subsubsection{Defining measures in-terms of information-leakages}
The privacy, interpretability, and transferability measures are defined in-terms of information-leakages:
\begin{itemize}
\item privacy-leakage is measured as the amount of information about private/sensitive variables leaked by the shared variables;
\item interpretability is measured as the amount of information about interpretable parameters leaked by the model;
\item transferability is measured as the amount of information about the source domain model output leaked by the target domain model output.
\end{itemize}
\subsubsection{Variational membership-mapping Bayesian models}
In order to derive analytical expressions for the defined privacy-leakage, interpretability, and transferability measures, the stochastic inverse models (governing the relationships amongst variables) will be required. In this study, we leverage the variational membership-mapping learning solution to build the required stochastic inverse models. Membership-mappings~\cite{10.1007/978-3-030-87101-7_13,10.1007/978-3-030-87101-7_14} have been introduced as alternative to deep neural networks to address the issues such as determining the optimal model structure, smaller training dataset, and iterative time-consuming nature of numerical learning algorithms~\cite{8888203,9216097,KUMAR20211}. A membership-mapping represents data through a fuzzy set with a membership function such that the dimension of membership function increases with an increasing data size. A remarkable feature of membership-mappings is to allow an analytical approach to the variational learning of a membership-mappings based data representation model. Our idea is to employ membership-mappings for defining a stochastic inverse model which is inferred using variational Bayesian methodology.             
\subsubsection{Variational approximation of information theoretic measures} 
The variational membership-mapping Bayesian models are used to determine the lower bounds on the defined information theoretic measures for privacy-leakage, interpretability, and transferability. The lower bounds on measures are maximized using variational optimization methodology to derive analytically the expressions for approximating the  privacy-leakage, interpretability, and transferability measures. The analytically derived expressions form the basis for developing an algorithm for practically computing the measures using available data samples, where expectations over unknown distributions are approximated via sample-averages.      
\subsection{Novelty and Contributions}
This study demonstrates the proposed ITTAI framework via considering a unified approach to ``privacy-preserving interpretable and transferable learning'', which is the novelty of this study. Further, the study introduces the novel information theoretic measures for privacy-leakage, interpretability, and transferability. It is possible to derive analytical expressions for the defined measures, provided a knowledge regarding the statistical data distributions is available. However, in practice, the data distributions are unknown and thus a way to approximate the defined measures is required. Therefore, a novel method, that employs recently introduced membership-mappings~\cite{10.1007/978-3-030-87101-7_13,10.1007/978-3-030-87101-7_14,8888203,9216097,KUMAR20211}, is presented for approximating the defined privacy-leakage, interpretability, and transferability measures. The method relies on inferring a variational Bayesian model that facilitates an analytical approximation of the information theoretic measures through variational optimization methodology. A computational algorithm is provided for practically calculating the privacy-leakage, interpretability, and transferability measures. Finally, an algorithm is presented that provides 
\begin{enumerate}
\item information theoretic evaluation of privacy-leakage, interpretability, and transferability in a semi-supervised transfer and multi-task learning scenario;
\item an adversary model for estimating private data and thus for simulating privacy attacks;
\item an interpretability model for estimating interpretable parameters and thus for providing an interpretation to the non-interpretable data vectors.
\end{enumerate}       
To the best knowledge of the authors, no previous study presented a unified information theoretic approach to study different aspects of trustworthy AI in a rigorous analytical manner. This is the main contribution of this text.  
\subsection{Organization}
This text is organized into sections. The proposed methodology in this study relies on the membership-mappings for data representation learning. Therefore, section~\ref{sec_2} has been dedicated to the review of membership-mappings based transferrable learning methodology. An application of membership-mappings to solve an inverse modeling problem via developing a variational membership-mapping Bayesian model is considered in section~\ref{section_738539.8242}. Section~\ref{sec_738541.6948} presents the most important result of this study regarding variational approximation of information-leakage and development of a computational algorithm for calculating information-leakage. The significance of information-leakage evaluation is due to the measures (for privacy-leakage, interpretability, and transferability) which are formally introduced in section~\ref{sec_measures}. Section~\ref{sec_measures} further provides an algorithm to study the privacy, interpretability, and transferability aspects in a unified manner. The application of proposed measures to study the tradeoffs is also demonstrated through the experiments made on the widely used MNIST and ``Office+Caltech256'' datasets in section~\ref{sec_experiments}. Section~\ref{sec_experiments} further considers a biomedical application concerned with the detection of mental stress on individual using heart rate variability analysis. Finally, the concluding remarks are provided in section~\ref{sec_conclusion}.

\section{Mathematical Background}\label{sec_2}
This section reviews the membership-mappings and transferable deep learning from~\cite{10.1007/978-3-030-87101-7_13,10.1007/978-3-030-87101-7_14,kumar2022differentially}
\subsection{Notations}
\begin{itemize}
\item Let $n,N,p,M \in \mathbb{N}$. 
\item Let $\mathcal{B}(\mathbb{R}^N)$ denote the \emph{Borel $\sigma-$algebra} on $\mathbb{R}^N$, and let $\lambda^N$ denote the \emph{Lebesgue measure} on $\mathcal{B}(\mathbb{R}^N)$. 
\item Let $(\mathcal{X}, \mathcal{A} , \rho)$ be a probability space with unknown probability measure $\rho$. 
\item Let us denote by $\mathcal{S}$ the set of finite samples of data points drawn i.i.d. from $\rho$, i.e.,
\begin{IEEEeqnarray}{rCl}  
\mathcal{S} & := &  \{ (x^i  \sim \rho )_{i=1}^N \; | \; N \in \mathbb{N} \}.
\end{IEEEeqnarray} 
\item For a sequence $\mathrm{x} = (x^1,\cdots,x^N) \in \mathcal{S}$, let $|\mathrm{x}|$ denote the cardinality i.e. $|\mathrm{x}| = N$.     
\item If $\mathrm{x} = (x^1, \cdots, x^N),\; \mathrm{a}= (a^1, \cdots, a^M) \in \mathcal{S}$, then $\mathrm{x} \wedge \mathrm{a}$ denotes the concatenation of the sequences $\mathrm{x}$ and $\mathrm{a}$, i.e., $\mathrm{x} \wedge \mathrm{a} = (x^1, \ldots, x^N, a^1, \ldots, a^M)$.
\item Let us denote by $\mathbb{F}(\mathcal{X})$ the set of $\mathcal{A}$-$\mathcal{B}(\mathbb{R})$ measurable functions $f:\mathcal{X} \rightarrow \mathbb{R}$, i.e.,
\begin{IEEEeqnarray}{rCl}  
\mathbb{F}(\mathcal{X}) & := &  \{ f:\mathcal{X} \rightarrow \mathbb{R}  \; | \;  \mbox{$f$ is $\mathcal{A}$-$\mathcal{B}(\mathbb{R})$ measurable}\}.
\end{IEEEeqnarray} 
\item For convenience, the values of a function $f \in \mathbb{F}(\mathcal{X})$ at points in the collection $\mathrm{x} = (x^1,\cdots,x^N)$ are represented as $f(\mathrm{x})=(f(x^1),\cdots,f(x^N))$.
\item For a given $\mathrm{x} \in \mathcal{S}$ and $A\in \mathcal{B}(\mathbb{R}^{|\mathrm{x}|})$, the cylinder set $\mathcal{T}_{\mathrm{x}}(A)$ in $\mathbb{F}(\mathcal{X})$ is defined as 
\begin{IEEEeqnarray}{rCl}
\mathcal{T}_{\mathrm{x}}(A) &: = & \{ f \in \mathbb{F}(\mathcal{X}) \; | \; f(\mathrm{x}) \in A   \}.
\end{IEEEeqnarray}
\item Let $\mathcal{T}$ be the family of cylinder sets defined as
\begin{IEEEeqnarray}{rCl}
\mathcal{T} & := & \left\{ \mathcal{T}_{\mathrm{x}}(A)\; | \; A \in \mathcal{B}(\mathbb{R}^{|\mathrm{x}|}),\; \mathrm{x} \in \mathcal{S} \right \}.
\end{IEEEeqnarray} 
\item Let $\sigma(\mathcal{T})$ be the $\sigma$-algebra generated by $\mathcal{T}$.
\item Given two $\mathcal{B}(\mathbb{R}^N)-\mathcal{B}(\mathbb{R})$ measurable mappings, $g:\mathbb{R}^N \rightarrow \mathbb{R}$ and $\mu:\mathbb{R}^N \rightarrow \mathbb{R}$, the weighted average of $g(\mathrm{y})$ over all $\mathrm{y} \in \mathbb{R}^{N}$, with $\mu(\mathrm{y})$ as the weighting function, is computed as   
\begin{IEEEeqnarray}{rCl}
\label{eq_738118.427179} \left< g \right>_{\mu}& := & \frac{1}{ \int_{\mathbb{R}^{N}}  \mu(\mathrm{y})\, \dd\lambda^{N}(\mathrm{y})} \int_{\mathbb{R}^{N}} g(\mathrm{y}) \mu(\mathrm{y})\, \dd \lambda^{N}(\mathrm{y}).
\end{IEEEeqnarray} 
\item Let $\zeta_{\mathrm{x}}:\mathbb{R}^{|\mathrm{x}|} \rightarrow [0,1]$ be a membership function satisfying the following properties:
\begin{description}
    \item[Nowhere Vanishing:] $\zeta_{\mathrm{x}}(\mathrm{y}) > 0$ for all $\mathrm{y} \in \mathbb{R}^{|\mathrm{x}|}$, i.e.,
\begin{IEEEeqnarray}{rCl}
    \label{eq:supp}
     \mbox{supp}[\zeta_{\mathrm{x}}] & = & \mathbb{R}^{|\mathrm{x}|}.
\end{IEEEeqnarray}   
    \item[Positive and Bounded Integrals:] the functions $\zeta_{\mathrm{x}}$ are absolutely continuous and Lebesgue integrable over the whole domain such that for all $\mathrm{x}\in \mathcal{S}$ we have
     \begin{eqnarray}
    \label{eq:positive}
   0 < \int_{\mathbb{R}^{|\mathrm{x}|}} \zeta_{\mathrm{x}}\, \dd \lambda^{|\mathrm{x}|}  < \infty.
   \end{eqnarray}
   \item[Consistency of Induced Probability Measure:] the membership function induced probability measures $\mathbb{P}_{\zeta_{\mathrm{x}}}$, defined on any $A \in \mathcal{B}(\mathbb{R}^{|\mathrm{x}|})$, as
\begin{IEEEeqnarray}{rCl}
\mathbb{P}_{\zeta_{\mathrm{x}}}(A) & := &  \frac{1}{ \int_{\mathbb{R}^{|\mathrm{x}|}} \zeta_{\mathrm{x}}\, \dd \lambda^{|\mathrm{x}|}} \int_{A} \zeta_{\mathrm{x}}\, \dd\lambda^{|\mathrm{x}|}
\end{IEEEeqnarray}  
are consistent in the sense that for all $\mathrm{x},\;\mathrm{a} \in \mathcal{S}$:
\begin{IEEEeqnarray}{rCl}
\label{eq_738083.390026} \mathbb{P}_{\zeta_{\mathrm{x} \wedge \mathrm{a}}}(A \times \mathbb{R}^{|\mathrm{a}|}) & = & \mathbb{P}_{\zeta_{\mathrm{x}}}(A). 
\end{IEEEeqnarray}  
\end{description}
The collection of membership functions satisfying aforementioned assumptions is denoted by 
\begin{IEEEeqnarray}{rCl}
\Theta & := & \{ \zeta_{\mathrm{x}}:\mathbb{R}^{|\mathrm{x}|} \rightarrow [0,1] \; | \; (\ref{eq:supp}),  (\ref{eq:positive}), (\ref{eq_738083.390026}),\; \mathrm{x} \in \mathcal{S}\}.
\end{IEEEeqnarray}
\end{itemize}
\subsection{Review of Variational Membership-Mappings}
\begin{definition}[Student-t Membership-Mapping~\cite{10.1007/978-3-030-87101-7_13}]\label{def_student_t_set_membership_mapping}
A Student-t membership-mapping, $\mathcal{F} \in \mathbb{F}(\mathcal{X})$, is a mapping with input space $\mathcal{X} = \mathbb{R}^n$ and a membership function $\zeta_{\mathrm{x}} \in \Theta$ that is Student-t like:
\begin{IEEEeqnarray}{rCl}
\label{eq_student_t_membership} 
\label{eq_738098.751419}\zeta_{\mathrm{x}}(\mathrm{y}) & = & \left(1 + 1/(\nu - 2) \left( \mathrm{y} - \mathrm{m}_{\mathrm{y}} \right)^T K^{-1}_{\mathrm{x}\mathrm{x}} \left( \mathrm{y}- \mathrm{m}_{\mathrm{y}}\right) \right)^{-\frac{\nu+|\mathrm{x}|}{2}}
\end{IEEEeqnarray} 
where $\mathrm{x} \in \mathcal{S}$, $\mathrm{y} \in \mathbb{R}^{|\mathrm{x}|}$, $\nu \in \mathbb{R}_{+}\setminus [0,2]$ is the degrees of freedom, $\mathrm{m}_{\mathrm{y}} \in \mathbb{R}^{|\mathrm{x}|}$ is the mean vector, and $K_{\mathrm{x}\mathrm{x}} \in \mathbb{R}^{|\mathrm{x}| \times |\mathrm{x}|}$ is the covariance matrix with its $(i,j)-$th element given as 
\begin{IEEEeqnarray}{rCl}
\label{738026.844153}  (K_{\mathrm{x}\mathrm{x}})_{i,j} & = & kr(x^i,x^j) 
 \end{IEEEeqnarray}  
where $kr: \mathbb{R}^n \times \mathbb{R}^n \rightarrow \mathbb{R}$ is a positive definite kernel function defined as 
\begin{IEEEeqnarray}{rCl}
\label{eq_membership1003_3} kr(x^{i},x^{j}) & = &  \sigma^2 \exp \left(-0.5\sum_{k = 1}^{n} w_{k} \left |  x^{i}_k - x^{j}_k \right |^2\right)
 \end{IEEEeqnarray}  
where $x_k^i$ is the $k-$th element of $x^i$, $\sigma^2$ is the variance parameter, and $w_{k} \geq 0$ (for $k \in \{1,\cdots,n\}$).
\end{definition}
Given a dataset $\{(x^i,y^i)\;|\; x^i \in \mathbb{R}^n,\;y^i \in \mathbb{R}^p,\; i \in \{1,\cdots,N \} \}$, it is assumed that there exist zero-mean Student-t membership-mappings $\mathcal{F}_1, \cdots, \mathcal{F}_p \in \mathbb{F}(\mathbb{R}^n)$ such that
\begin{IEEEeqnarray}{rCl}
\label{eq_738118.641846} y^i &\approx & \left[\begin{IEEEeqnarraybox*}[][c]{,c/c/c,} \mathcal{F}_1(x^i)  & \cdots & \mathcal{F}_p(x^i) \end{IEEEeqnarraybox*} \right]^T.
\end{IEEEeqnarray}   
Under modeling scenario~(\ref{eq_738118.641846}), \cite{kumar2022differentially} presents an algorithm (stated as Algorithm~\ref{algorithm_basic_learning} in Appendix~\ref{appendix_algorithms}) for the variational learning of membership-mappings.  
\begin{definition}[Membership-Mappings Prediction~\cite{kumar2022differentially}]
Given the parameters set $\mathbb{M} = \{\alpha, \mathrm{a}, M,\sigma,w\}$ returned by Algorithm~\ref{algorithm_basic_learning}, the learned membership-mappings could be used to predict output corresponding to any arbitrary input data point $x \in \mathbb{R}^n$ as
\begin{IEEEeqnarray}{rCl}
\label{eq_738124.770095}\hat{y}(x;\mathbb{M}) & = & \alpha^T(G(x))^T
\end{IEEEeqnarray}
where $G(\cdot) \in \mathbb{R}^{1 \times M}$ is a vector-valued function~(\ref{eq_738495.5497}). 
\end{definition}
\subsection{Review of Membership-Mappings Based Conditionally Deep Autoencoders}
\begin{definition}[Membership-Mapping Autoencoder~\cite{10.1007/978-3-030-87101-7_14}]\label{def_SFMA}
A membership-mapping autoencoder, $\mathcal{G}:\mathbb{R}^p \rightarrow \mathbb{R}^p$, maps an input vector $y \in \mathbb{R}^p$ to $\mathcal{G}(y) \in \mathbb{R}^p$ such that 
 \begin{IEEEeqnarray}{rCl}
\label{eq_added_before_publication_1}  \mathcal{G}(y) &    \overset{\underset{\mathrm{def}}{}}{=} &  \left[\begin{IEEEeqnarraybox*}[][c]{,c/c/c,}  \mathcal{F}_1(Py) & \cdots &  \mathcal{F}_p(Py)
 \end{IEEEeqnarraybox*} \right]^T, 
\end{IEEEeqnarray} 
where $\mathcal{F}_j$ ($j \in \{1,2,\cdots,p\}$) is a Student-t membership-mapping, $P \in \mathbb{R}^{n \times p} (n \leq p)$ is a matrix such that the product $Py$ is a lower-dimensional encoding for $y$. 
\end{definition}
\begin{definition}[Conditionally Deep Membership-Mapping Autoencoder (CDMMA)~\cite{10.1007/978-3-030-87101-7_14,kumar2022differentially}]\label{def_deep_autoencoder}
A conditionally deep membership-mapping autoencoder, $\mathcal{D}:\mathbb{R}^p \rightarrow \mathbb{R}^p$, maps a vector $y \in \mathbb{R}^p$ to $\mathcal{D}(y) \in \mathbb{R}^p$ through a nested composition of finite number of membership-mapping autoencoders such that
 \begin{IEEEeqnarray}{rCl}
 y^l & = & (\mathcal{G}_l \circ \cdots \circ \mathcal{G}_2\circ \mathcal{G}_1)(y), \; \forall l \in \{1,2,\cdots,L \}\\
 l^* & = & \arg\;\min_{l \: {\in} \: \{1,2,\cdots,L \}}\; \| y -  y^l \|^2 \\
  \mathcal{D}(y) & = & y^{l^*},
\end{IEEEeqnarray}  
where $\mathcal{G}_l(\cdot)$ is a membership-mapping autoencoder (Definition~\ref{def_SFMA}).
\end{definition}
An algorithm (stated as Algorithm~\ref{algorithm_DSFMA} in Appendix~\ref{appendix_algorithms}) has been provided in \cite{kumar2022differentially} for the variational learning of CDMMA.
\begin{definition}[CDMMA Filtering~\cite{10.1007/978-3-030-87101-7_14,kumar2022differentially}]\label{def_DSFMA_filtering}
Given a CDMMA with its parameters being represented by a set $\mathcal{M} = \{\{\mathbb{M}^1,\cdots,\mathbb{M}^L\}, \{P^1,\cdots,P^L \} \}$, the autoencoder can be applied for filtering a given input vector $y \in \mathbb{R}^p$ as follows:   
 \begin{IEEEeqnarray}{rCl}
x^l(y;\mathcal{M}) &=& \left\{ \,
    \begin{IEEEeqnarraybox}[][c]{l?s}
      \IEEEstrut
      P^ly, & $l=1$ \\
      P^l  \hat{y}^{l-1}(x^{l-1};\mathbb{M}^{l-1})&  $l \geq 2$
      \IEEEstrut
    \end{IEEEeqnarraybox}
\right. 
\end{IEEEeqnarray} 
Here, $\hat{y}^{l-1}$ is the output of the $(l-1)-$th layer estimated using (\ref{eq_738124.770095}). Finally, CDMMA's output, $\mathcal{D}(y;\mathcal{M})$, is given as
\begin{IEEEeqnarray}{rCl}
\label{eq_satguru_18}  \widehat{\mathcal{D}}(y;\mathcal{M}) & = &  \hat{y}^{l^*}(x^{l^*};\mathbb{M}^{l^*}) \\
\label{eq_satguru_19} l^*  & = & \arg\;\min_{l \: {\in} \: \{1,\cdots,L \}}\; \|y - \hat{y}^{l}(x^{l};\mathbb{M}^{l}) \|^2.
 \end{IEEEeqnarray} 
\end{definition}
\begin{definition}[A Wide CDMMA~\cite{10.1007/978-3-030-87101-7_14,kumar2022differentially}]\label{def_wide_deep_autoencoder}
A wide CDMMA, $\mathcal{WD}:\mathbb{R}^p \rightarrow \mathbb{R}^p$, maps a vector $y \in \mathbb{R}^p$ to $\mathcal{WD}(y) \in \mathbb{R}^p$ through a parallel composition of $S$  ($S \in \mathcal{Z}_+$) number of CDMMAs such that
 \begin{IEEEeqnarray}{rCl}
\label{eq_738125.489500} \mathcal{WD}(y) & = & \mathcal{D}_{s^*}(y)\\
 s^* & = & \arg\;\min_{s \in \{1,2,\cdots,S \}}\; \| y -  \mathcal{D}_s(y)  \|^2,
 \end{IEEEeqnarray}  
where $\mathcal{D}_s(y)$ is the output of $s-$th CDMMA. 
\end{definition}
Algorithm~\ref{algorithm_WDSFMA} (in Appendix~\ref{appendix_algorithms}) follows from~\cite{kumar2022differentially} for the variational learning of wide CDMMA. 
\begin{definition}[Wide CDMMA Filtering~\cite{10.1007/978-3-030-87101-7_14,kumar2022differentially}]\label{def_DSFMA_filtering}
Given a wide CDMMA with its parameters being represented by a set $\mathcal{P} = \{\mathcal{M}^s\}_{s=1}^S$, the autoencoder can be applied for filtering a given input vector $y \in \mathbb{R}^p$ as follows:   
  \begin{IEEEeqnarray}{rCl}
 \label{eq_738500.4495}\widehat{\mathcal{WD}}(y;\mathcal{P}) & = &   \widehat{\mathcal{D}}(y;\mathcal{M}^{s^*})\\
 s^* & = & \arg\;\min_{s \in \{1,2,\cdots,S \}}\; \| y -  \widehat{\mathcal{D}}(y;\mathcal{M}^{s})  \|^2,
 \end{IEEEeqnarray}   
where $ \widehat{\mathcal{D}}(y;\mathcal{M}^{s})$ is the output of $s-$th CDMMA estimated using (\ref{eq_satguru_18}).
\end{definition}
\subsection{Membership-Mappings for Classification}
A classifier (i.e. Definition~\ref{def_classifier}) and an algorithm for its variational learning (stated as Algorithm~\ref{algorithm_classification} in Appendix~\ref{appendix_algorithms}) follows from~\cite{10.1007/978-3-030-87101-7_14,kumar2022differentially}.
\begin{definition}[A Classifier~\cite{10.1007/978-3-030-87101-7_14,kumar2022differentially}]\label{def_classifier}
A classifier, $\mathcal{C}: \mathbb{R}^p \rightarrow \{1,2,\cdots, C\}$, maps a vector $y \in \mathbb{R}^p$ to $\mathcal{C}(y) \in  \{1,2,\cdots, C\}$ such that
\begin{IEEEeqnarray}{rCl}
\label{eq_satguru_20} \mathcal{C}(y; \{\mathcal{P}_c\}_{c=1}^C)
  & = &  \arg\;\min_{c \:{\in}\: \{1,2,\cdots,C \}}\; \| y -   \widehat{\mathcal{WD}}(y;\mathcal{P}_c) \|^2  \end{IEEEeqnarray} 
where $ \widehat{\mathcal{WD}}(y;\mathcal{P}_c)$, computed using (\ref{eq_738500.4495}), is the output of $c-$th wide CDMMA. The classifier assigns to an input vector the label of that class whose associated autoencoder best reconstructs the input vector.      
 \end{definition}
\subsection{Review of Membership-Mappings Based Privacy-Preserving Transferrable Learning}  
Privacy-preserving semi-supervised transfer and multi-task learning problem has been recently addressed in \cite{kumar2022differentially} by means of variational membership-mappings. The method, as suggested in \cite{kumar2022differentially}, involves the following steps:  
\paragraph{Optimal noise adding mechanism for differentially private classifiers:} 
 The approach suggested in~\cite{kumar2022differentially} relies on a tailored noise adding mechanism to achieve a given level of differential privacy-loss bound with the minimum perturbation of the data. In particularly, Algorithm~\ref{algorithm_differential_private_approximation} (in Appendix~\ref{appendix_algorithms}) is suggested for a differentially private approximation of data samples and Algorithm~\ref{algorithm_private_classification} (in Appendix~\ref{appendix_algorithms}) is suggested for building a differentially private classifier.    
\paragraph{Semi-supervised transfer learning scenario:} 
The aim is to transfer the knowledge extracted by a classifier trained using source dataset to the classifier of target domain such that privacy of source dataset is preserved. Let $\{\mathbf{Y}^{sr}_c \}_{c=1}^C$ be the labelled source dataset where $\mathbf{Y}^{sr}_c = \{ y^{i,c}_{sr} \in \mathbb{R}^{p_{sr}} \; | \; i \in \{1,\cdots,N^{sr}_c \} \}$ represents $c-$th labelled samples. The target dataset consist of a few labelled samples $\{\mathbf{Y}^{tg}_c \}_{c=1}^C$ (with $\mathbf{Y}^{tg}_c = \{ y^{i,c}_{tg} \in \mathbb{R}^{p_{tg}} \; | \; i \in \{1,\cdots,N^{tg}_c \} \}$) and another set of unlabelled samples $\mathbf{Y}^{tg}_* = \{ y^{i,*}_{tg} \in \mathbb{R}^{p_{tg}} \; | \; i \in \{1,\cdots,N^{tg}_* \} \}$. 
\paragraph{Differentially private source domain classifier:} 
For a given differential privacy parameters: $d,\epsilon,\delta$; Algorithm~\ref{algorithm_differential_private_approximation} (in Appendix~\ref{appendix_algorithms}) is applied on $\mathbf{Y}^{sr}_c$ to obtain the differentially private approximated data samples, $\mathbf{Y}^{+sr}_c = \{ y^{+i,c}_{sr} \in \mathbb{R}^{p_{sr}} \; | \; i \in \{1,\cdots,N^{sr}_c \}\}$, for all $c \in \{1,\cdots,C \}$. Algorithm~\ref{algorithm_private_classification} (in Appendix~\ref{appendix_algorithms}) is applied on $\{\mathbf{Y}^{+sr}_c \}_{c=1}^C$ to build a differentially private source domain classifier characterized by parameters sets $ \{ \mathcal{P}_c^{+sr}  \}_{c=1}^C$.  
\paragraph{Latent subspace transformation-matrices:}
For a given subspace dimension $n_{st}  \in \{1,2,\cdots,\min(p_{sr},p_{tg})\}$, the source domain transformation-matrix $V^{+sr} \in \mathbb{R}^{n_{st}  \times p_{sr}}$ is defined as with its $i-$th row equal to transpose of eigenvector corresponding to $i-$th largest eigenvalue of sample covariance matrix computed on differentially private approximated source samples. The target domain transformation-matrix $V^{tg} \in \mathbb{R}^{n_{st} \times p_{tg}}$ is defined as with its $i-$th row equal to transpose of eigenvector corresponding to $i-$th largest eigenvalue of sample covariance matrix computed on target samples. 
\paragraph{Subspace alignment:}
A target sample is mapped to source-data-space via following transformation: 
\begin{IEEEeqnarray}{CCl}
\label{eq_1643012238} y_{tg \rightarrow sr}(y_{tg}) &  = & \left \{\begin{array}{ll}  y_{tg}, & p_{sr} = p_{tg} \\
(V^{+sr})^TV^{tg}y_{tg}, & p_{sr} \neq p_{tg}
\end{array} \right. 
\end{IEEEeqnarray}   
Both labelled and unlabelled target datasets are transformed to define the following sets: 
\begin{IEEEeqnarray}{CCl}
\label{eq_738522.553647}\mathbf{Y}^{tg \rightarrow sr}_c & := & \{ y_{tg \rightarrow sr}(y_{tg})   \; | \; y_{tg} \in \mathbf{Y}^{tg}_c \} \\
\label{eq_738522.553880} \mathbf{Y}^{tg \rightarrow sr}_* & := & \{ y_{tg \rightarrow sr}(y_{tg}) \; | \; y_{tg} \in \mathbf{Y}^{tg}_* \}.
\end{IEEEeqnarray}   
 \paragraph{Target domain classifier:} 
The $k-$th iteration for building the target domain classifier, where $k \in \{1,\cdots,it\_{max} \}$, consists of following updates:  
  \begin{IEEEeqnarray}{CCl}
    \label{eq_iterative_target_classification}  \{ \mathcal{P}_c^{tg}|_k  \}_{c=1}^C
 & = &   \text{Algorithm~\ref{algorithm_classification}}\left( \left \{\mathbf{Y}^{tg \rightarrow sr}_{c} \cup \mathbf{Y}^{tg \rightarrow sr}_{*,c}|_{k-1} \right \}_{c=1}^C, n|_k, r_{max},L \right) \\
  \label{eq_iterative_target_samples_class_c} \mathbf{Y}^{tg \rightarrow sr}_{*,c}|_{k} & = & \left \{ y^{i,*}_{tg \rightarrow sr} \in \mathbf{Y}^{tg \rightarrow sr}_*  \; | \; \mathcal{C}(y^{i,*}_{tg \rightarrow sr};\{ \mathcal{P}_c^{tg}|_{k} \}_{c=1}^C)  = c,\;  i \in \{1,\cdots,N^{tg}_* \} \right \}
  \end{IEEEeqnarray}  
where $\left\{n|_1,n|_2,\cdots \right \}$ is a monotonically non-decreasing sequence. 
\paragraph{source2target model:}
The mapping from source to target domain is learned by means of a variational membership-mappings based model as in the following:     \begin{IEEEeqnarray}{CCl}
 \label{eq_738522.588899} \mathbb{M}^{sr\rightarrow tg}& = &    \text{Algorithm~\ref{algorithm_basic_learning}}\left(\mathcal{D}, M_{max} \right) \\
 \label{eq_738522.588685}\mathcal{D}   &:= & \left \{ \left( \widehat{\mathcal{WD}}(y;\mathcal{P}_c^{+sr}), y \right)  \; | \; y \in \left \{ \mathbf{Y}^{tg \rightarrow sr}_{c} \cup \mathbf{Y}^{tg \rightarrow sr}_{*,c}|_{it\_{max}} \right \},\; c \in \left \{1,\cdots,C  \right \} \right \} \\
 M_{max} & = & \min(\lceil N^{tg}/2 \rceil,1000)
   \end{IEEEeqnarray} 
where $N^{tg} = | \mathcal{D} |$ is the total number of target samples, $\widehat{\mathcal{WD}}(\cdot;\cdot)$ is defined as in~(\ref{eq_738500.4495}), $\mathbf{Y}^{tg \rightarrow sr}_{c}$ is defined as in~(\ref{eq_738522.553647}), and $\mathbf{Y}^{tg \rightarrow sr}_{*,c}$ is defined as in (\ref{eq_iterative_target_samples_class_c}). 
\paragraph{Transfer and multi-task learning:}
Both source and target domain classifiers are combined with source2target model for predicting the label associated to a target sample $y_{tg \rightarrow sr}$ as 
  \begin{IEEEeqnarray}{CCl}
\nonumber  \hat{c}(y_{tg \rightarrow sr};\{ \mathcal{P}_c^{tg} \}_{c=1}^C, \{ \mathcal{P}_c^{+sr} \}_{c=1}^C,  \mathbb{M}^{sr\rightarrow tg})  & = & \arg\;\min_{c \:{\in}\: \{1,2,\cdots,C \}}\;  \left \{ \min\left(\left \| y_{tg \rightarrow sr} - \widehat{\mathcal{WD}}(y_{tg \rightarrow sr};\mathcal{P}_c^{tg})  \right \|^2, \right. \right. \\
\nonumber  && \left. \left. \left \| y_{tg \rightarrow sr}  - \hat{y} \left( \widehat{\mathcal{WD}}(y_{tg \rightarrow sr};\mathcal{P}_c^{+sr}) ; \mathbb{M}^{sr\rightarrow tg} \right) \right \|^2, \right. \right. \\
\label{eq_predicted_target_label_multitask} && \left. \left. \left \| y_{tg \rightarrow sr} - \widehat{\mathcal{WD}}(y_{tg \rightarrow sr};\mathcal{P}_c^{+sr})  \right \|^2   \right) \right \}. \IEEEeqnarraynumspace
\end{IEEEeqnarray}
where $\hat{y} \left( \cdot ; \mathbb{M}^{sr\rightarrow tg} \right)$ is the output of source2target model computed using (\ref{eq_738124.770095}). 
\section{Variational Membership-Mapping Bayesian Models}\label{section_738539.8242}
We consider the application of membership-mappings to solve the inverse modeling problem related to $x = f_{t \rightarrow x}(t)$, where $f_{t \rightarrow x}: \mathbb{R}^q \rightarrow \mathbb{R}^n$ is a forward map. Specifically, a membership-mappings model is used to approximate the inverse mapping $f_{t \rightarrow x}^{-1}$. 
\subsection{A Prior Model}
Given a dataset: $\{(x^i,t^i) \; | \; i \in \{1,\cdots,N \} \}$, Algorithm~\ref{algorithm_basic_learning} can be used to build a membership-mappings model characterized by a set of parameters, say $\mathbb{M}^{x \rightarrow t} = \{\alpha^{x \rightarrow t}, \mathrm{a}, M,\sigma,w\}$ (where $x \rightarrow t$ indicates the mapping from $x$ to $t$ has been approximated by the membership-mappings). It follows from~(\ref{eq_738124.770095}) that the membership-mappings model predicted output corresponding to an input $x$ is given as 
\begin{IEEEeqnarray}{rCl}
\label{eq_738540.6889}\hat{t}(x;\mathbb{M}^{x \rightarrow t}) & = & (\alpha^{x \rightarrow t})^T(G(x))^T
\end{IEEEeqnarray}
where $G(\cdot) \in \mathbb{R}^{1 \times M}$ is a vector-valued function defined as in~(\ref{eq_738495.5497}). The $k-$th element of $\hat{t}$ is given as
\begin{IEEEeqnarray}{rCl}
\label{eq_738541.429207}\hat{t}_k(x;\mathbb{M}^{x \rightarrow t}) & = & (G(x))\alpha^{x \rightarrow t}_k
\end{IEEEeqnarray}
where $\alpha^{x \rightarrow t}_k$ is $k-$th column of matrix $\alpha^{x \rightarrow t}$. 

Expression~(\ref{eq_738541.429207}) allows to estimate for any arbitrary $x$ the corresponding $t$ using membership-mappings model. This motivates introducing the following prior model:        
 \begin{IEEEeqnarray}{rCl}
\label{eq_738130.729430} t_k & = &  \left(G(x) \right) \theta_k + e_k\\
\label{eq_738130.729747} \theta_k & \sim & \mathcal{N}(\alpha^{x \rightarrow t}_k,\Lambda_k^{-1}) \\
\label{eq_738130.729972} e_k & \sim & \mathcal{N}(0,\gamma^{-1})\\
\label{eq_738130.730170} \gamma & \sim & \text{Gamma}(a_{\gamma},b_{\gamma})
\end{IEEEeqnarray}         
where $k \in \{1,\cdots,q\}$; $ \mathcal{N}(\alpha^{x \rightarrow t}_k,\Lambda_k^{-1}) $ is the multivariate normal distribution with mean $\alpha^{x \rightarrow t}_k$ and covariance $\Lambda_k^{-1}$; and $\text{Gamma}(a_{\gamma},b_{\gamma})$ is the Gamma distribution with shape parameter $a_{\gamma}$ and rate parameter $b_{\gamma}$. The estimation provided by membership-mappings model $\mathbb{M}^{x \rightarrow t}$ (i.e. (\ref{eq_738541.429207})) is incorporated by the prior model~(\ref{eq_738130.729430}-\ref{eq_738130.730170}), since
 \begin{IEEEeqnarray}{rCl}
 \mathbb{E}[t_k] & = & \hat{t}_k(x;\mathbb{M}^{x \rightarrow t}). 
\end{IEEEeqnarray}                
\subsection{Variational Bayesian Inference}
Given the dataset, $\{ (x^i \in \mathbb{R}^n,t^i \in \mathbb{R}^q)\; | \; i \in \{1,2,\cdots,N \} \}$, the variational Bayesian method is considered for an inference of the stochastic model~(\ref{eq_738130.729430}), with priors as (\ref{eq_738130.729747}), (\ref{eq_738130.729972}), and (\ref{eq_738130.730170}).  For all $i \in \{1,\cdots,N \}$ and $k \in \{1,\cdots,q \}$, we have
 \begin{IEEEeqnarray}{rCl}
 t_k^i & = & \left(G(x^i) \right)\theta_k + e_k^i,
\end{IEEEeqnarray}    
where $\theta_k  \sim  \mathcal{N}(\alpha^{x \rightarrow t}_k,\Lambda_k^{-1})$ and $e_k^i  \sim  \mathcal{N}(0,\gamma^{-1})$. Define $\boldsymbol{\mathrm{t}}_k \in \mathbb{R}^N$, $  \boldsymbol{\mathrm{e}}_k \in \mathbb{R}^N$, and $ R_{\mathrm{y}} \in \mathbb{R}^{N \times M}$ as
 \begin{IEEEeqnarray}{rCl}
 \boldsymbol{\mathrm{t}}_k & = & \left[\begin{IEEEeqnarraybox*}[][c]{,c/c/c,}  t^1_k & \cdots &t^N_k
 \end{IEEEeqnarraybox*} \right]^T \\
 \boldsymbol{\mathrm{e}}_k & = & \left[\begin{IEEEeqnarraybox*}[][c]{,c/c/c,}  e^1_k & \cdots &e^N_k
 \end{IEEEeqnarraybox*} \right]^T \\
 R_{\mathrm{x}} & = & \left[\begin{IEEEeqnarraybox*}[][c]{,c/c/c,}  \left(G(x^1)\right)^T & \cdots & \left(G(x^N)\right)^T
 \end{IEEEeqnarraybox*} \right]^T.
\end{IEEEeqnarray}           
For all $k \in \{1,\cdots,q \}$, we have
 \begin{IEEEeqnarray}{rCl}
 \boldsymbol{\mathrm{t}}_k & = &   R_{\mathrm{x}} \theta_k +  \boldsymbol{\mathrm{e}}_k \\
p(\theta_k ; \alpha^{x \rightarrow t}_k, \Lambda_k) & = & \frac{1}{\sqrt{(2\pi)^M | (\Lambda_k)^{-1}|}} \exp\left(-0.5(\theta_k  - \alpha^{x \rightarrow t}_k)^T \Lambda_k (\theta_k  -\alpha^{x \rightarrow t}_k) \right) \IEEEeqnarraynumspace \\
 p( \boldsymbol{\mathrm{e}}_k  ; \gamma) & = & \frac{1}{\sqrt{(2\pi)^N (\gamma)^{-N}}} \exp\left(-0.5 \gamma \| \boldsymbol{\mathrm{e}}_k  \|^2  \right)\\
  p(\gamma ;  a_{\gamma},b_{\gamma}) & = & \left(b_{\gamma}^{a_{\gamma}}/ \Gamma(a_{\gamma}) \right) (\gamma)^{a_{\gamma}-1} \exp(-b_{\gamma} \gamma). 
\end{IEEEeqnarray} 
Define the following sets:
 \begin{IEEEeqnarray}{rCl}
  \mathbf{t} & = & \{\boldsymbol{\mathrm{t}}_1,\cdots,\boldsymbol{\mathrm{t}}_q\}\\
\label{eq_738132.476753} \theta & = & \{\theta_1,\cdots,\theta_q\}  
 \end{IEEEeqnarray}   
and consider the marginal probability of data $\mathbf{t} $ which is given as  
 \begin{IEEEeqnarray}{rCl}
p(\mathbf{t} ) & = &  \int \dd \theta \dd \gamma \: p(\mathbf{t} ,  \theta , \gamma).
 \end{IEEEeqnarray}   
Let $q(\theta, \gamma)$ be an arbitrary distribution. The log marginal probability of $\mathbf{t}$ can be expressed as 
 \begin{IEEEeqnarray}{rCl}
 \log(p(\mathbf{t})) & = & \int \dd \theta \dd \gamma \: q(\theta , \gamma) \log(p(\mathbf{t}))  \\
\label{eq_738131.433344} & = & \int \dd \theta \dd \gamma \: q(\theta , \gamma) \log\left( \frac{p(\mathbf{t},  \theta , \gamma)}{q(\theta , \gamma)}\right) + \int \dd \theta \dd \gamma \: q(\theta , \gamma) \log\left( \frac{q(\theta, \gamma)}{p( \theta , \gamma | \mathbf{t} )}\right). \IEEEeqnarraynumspace
 \end{IEEEeqnarray}  
Define
  \begin{IEEEeqnarray}{rCl}
\mathcal{L}(q(\theta , \gamma),\mathbf{t}) &  :=&  \int \dd \theta \dd \gamma \: q(\theta, \gamma) \log\left( p(\mathbf{t},  \theta, \gamma) / q(\theta , \gamma) \right) \IEEEeqnarraynumspace
 \end{IEEEeqnarray}  
to express (\ref{eq_738131.433344}) as
  \begin{IEEEeqnarray}{rCl}
 \log(p(\mathbf{t})) & = &  \mathcal{L}(q(\theta , \gamma),\mathbf{t}) + \mathrm{KL}(q(\theta , \gamma)\| p(\theta , \gamma | \mathbf{t})) \IEEEeqnarraynumspace
 \end{IEEEeqnarray}
 where $\mathrm{KL}$ is the Kullback-Leibler divergence of $p(\theta , \gamma | \mathbf{t})$ from $q(\theta , \gamma)$ and $\mathcal{L}$, referred to as negative free energy, provides a lower bound on the the logarithmic evidence for the data.

The variational Bayesian approach minimizes the difference (in term of $\mathrm{KL}$ divergence) between variational and true posteriors via analytically maximizing negative free energy $\mathcal{L}$ over variational distributions. However, the analytical derivation requires the following widely used mean-field approximation:  
\begin{IEEEeqnarray}{rCl}
\label{eq_738132.721205} q(\theta , \gamma) & = & q(\theta) q(\gamma) \\
 \label{eq_738132.721515} & = & q(\theta_1) \cdots q(\theta_q) q(\gamma).
 \end{IEEEeqnarray}  
Applying the standard variational optimization technique (as in \cite{5447695,7100899,5759770,6127913,7222472,KUMAR202187,10.1007/978-3-030-59028-4_4}), it can be verified that the optimal variational distributions maximizing $\mathcal{L}$ are as follows: 
 \begin{IEEEeqnarray}{rCl} 
 q^*(\theta_k ) & = & \frac{1}{\sqrt{(2\pi)^M | (\hat{\Lambda}_k)^{-1}|}} \exp\left(-0.5(\theta_k  - \hat{\mathrm{m}}_k)^T \hat{\Lambda}_k (\theta_k  - \hat{\mathrm{m}}_k) \right) \\
  q^*(\gamma) & = & \left((\hat{b}_{\gamma})^{\hat{a}_{\gamma}}/ \Gamma(\hat{a}_{\gamma}) \right) (\gamma)^{\hat{a}_{\gamma}-1} \exp(-\hat{b}_{\gamma} \gamma)
    \end{IEEEeqnarray}    
where the parameters $(\hat{\Lambda}_k,\hat{\mathrm{m}}_k,\hat{a}_{\gamma},\hat{b}_{\gamma})$ satisfy the following:
  \begin{IEEEeqnarray}{rCl} 
\label{eq_738131.510900}  \hat{\Lambda}_k & = & \Lambda_k + \left(\hat{a}_{\gamma}/ \hat{b}_{\gamma}\right) (R_{\mathrm{x}}  )^T R_{\mathrm{x}}  \\
\label{eq_738131.511110}  \hat{\mathrm{m}}_k & = & ( \hat{\Lambda}_k)^{-1}\left( \Lambda_k  \alpha^{x \rightarrow t}_k + \left(\hat{a}_{\gamma}/ \hat{b}_{\gamma}\right)  (R_{\mathrm{x}}  )^T  \boldsymbol{\mathrm{t}}_k \right) \\
\label{eq_738131.511349}   \hat{a}_{\gamma} & = & a_{\gamma} + 0.5qN \\
\label{eq_738131.511528} \hat{b}_{\gamma} & = & b_{\gamma}  + 0.5 \sum_{k=1}^q \left\{ \|\boldsymbol{\mathrm{t}}_k - R_{\mathrm{x}}   \hat{\mathrm{m}}_k \|^2  + Tr\left( (  \hat{\Lambda}_k)^{-1}  (R_{\mathrm{x}} )^T R_{\mathrm{x}}   \right) \right\}. \IEEEeqnarraynumspace
    \end{IEEEeqnarray} 
Algorithm~\ref{algorithm_variational_Bayesian_learning} is suggested for variational Bayesian inference of the model. The optimal distributions determined using Algorithm~\ref{algorithm_variational_Bayesian_learning} define the so-called \emph{Variational Membership-Mapping Bayesian Model (VMMBM)} as stated in Remark~\ref{remark_VBIM}.     
\begin{algorithm}
\caption{Variational membership-mapping Bayesian model inference}
\label{algorithm_variational_Bayesian_learning}
\begin{algorithmic}[1]
\Require  Dataset $\left\{ (x^i \in \mathbb{R}^n,t^i \in \mathbb{R}^q) \; | \; i \in \{1,\cdots,N \} \right \}$ and maximum possible number of auxiliary points $M_{max} \in \mathbb{Z}_+$ with $M_{max} \leq N$.  
\State Apply Algorithm~\ref{algorithm_basic_learning} on the dataset to build a variational membership-mappings model $\mathbb{M}^{x \rightarrow t}  = \{\alpha^{x \rightarrow t}, \mathrm{a}, M,\sigma,w\}$. 
\State For all $k \in \{1,\cdots,q\}$, choose $\Lambda_k = 10^{-3}I_M$. 
\State Choose $a_{\gamma} = 10^{-3}$ and $b_{\gamma} = 10^{-3}$.
\State Initialise $\hat{a}_{\gamma} / \hat{b}_{\gamma} = 1$.
\Repeat
\State update $\{\hat{\Lambda}_k,\hat{\mathrm{m}}_k\;|\;k \in \{1,\cdots,q \} \},\hat{a}_{\gamma},\hat{b}_{\gamma}$ using (\ref{eq_738131.510900}), (\ref{eq_738131.511110}), (\ref{eq_738131.511349}), (\ref{eq_738131.511528}).
 \Until{convergence.}
 \State \Return the parameters set $\mathbb{BM}^{x \rightarrow t} =  \{\{\hat{\mathrm{m}}_k,\hat{\Lambda}_k\;|\;k \in \{1,\cdots,q \} \},\hat{a}_{\gamma},\hat{b}_{\gamma} \} $. 
\end{algorithmic} 
\end{algorithm} 
\begin{remark}[Variational Membership-Mapping Bayesian Model (VMMBM)]\label{remark_VBIM}
The inverse mapping, $f_{t \rightarrow x}^{-1}$, is approximated as
\begin{IEEEeqnarray}{rCl}
\label{eq_738132.598922}t_k & = & \left(G(x)\right)  \theta_k + e_k,\\
\label{eq_738132.599286}\theta_k & \sim &  \mathcal{N}(\hat{\mathrm{m}}_k, \hat{\Lambda}_k^{-1}) \\
\label{eq_738132.599466}e_k & \sim & \mathcal{N}(0,\gamma^{-1})\\
\label{eq_738132.599647} \gamma & \sim & \emph{Gamma}(\hat{a}_{\gamma},\hat{b}_{\gamma}) 
\end{IEEEeqnarray}
where $k \in \{1,\cdots,q\}$ and $(\hat{\mathrm{m}}_k,\hat{\Lambda}_k,\hat{a}_{\gamma},\hat{b}_{\gamma})$ are returned by Algorithm~\ref{algorithm_variational_Bayesian_learning}.
\end{remark} 
\begin{remark}[Estimation by VMMBM] \label{remark_prediction_VBIM}
Given any $x^*$, the variational membership-mapping Bayesian model $\mathbb{BM}^{x \rightarrow t}$ (returned by Algorithm~\ref{algorithm_variational_Bayesian_learning}) can be used to estimate corresponding $t^*$ (such that $x^* = f_{t \rightarrow x}(t^*)$) as  
\begin{IEEEeqnarray}{rCl}
\label{eq_738185.490274}\tilde{t}(x^*;\mathbb{BM}^{x \rightarrow t}) & = & \left[\begin{IEEEeqnarraybox*}[][c]{,c/c/c,}   \left(G(x)\right) \hat{\mathrm{m}}_1  & \cdots & \left(G(x)\right) \hat{\mathrm{m}}_q \end{IEEEeqnarraybox*} \right]^T.
\end{IEEEeqnarray}
\end{remark}
\section{Evaluation of Information-Leakage}\label{sec_738541.6948}
Consider a scenario that a variable $t$ is related to another variable $x$ through a mapping $f_{t \rightarrow x}$ such that $x = f_{t \rightarrow x}(t)$. The mutual information $I(t;x)$ measures the amount of information obtained about variable $t$ through observing variable $x$. Since $x = f_{t\rightarrow x}(t)$, the entropy $H(t)$ remains fixed independent of mapping $f_{t\rightarrow x}$ and thus the quantity $I(t;x)-  H(t)$ is a measure of the amount of information about $t$ leaked by the mapping $f_{t\rightarrow x}$. 
\begin{definition}[Information-Leakage]\label{def_Information_Leakage}
Under the scenario that $x = f_{t \rightarrow x}(t)$, a measure of the amount of information about $t$ leaked by the mapping $f_{t\rightarrow x}$ is defined as
 \begin{IEEEeqnarray}{rCl} 
 IL_{f_{t\rightarrow x}} & := & I(t; f_{t\rightarrow x}(t))-  H(t) \\
                                       & = & I(t; x)-  H(t). 
\end{IEEEeqnarray}
The quantity $IL_{f_{t\rightarrow x}}$ is referred to as \emph{information-leakage}. 
\end{definition}
This section is dedicated to answer the question: \emph{How to calculate without knowing data distributions the information-leakage?} 
\subsection{Variational Approximation of Information-Leakage}
The mutual information between $t$ and $x$ is given as
\begin{IEEEeqnarray}{rCl}
I(t;x)& = & H(t) - H(t|x) \\
         & = & H(t) + \int p(t,x)\log\left( p(t|x)\right) \dd{t} \dd{x}\\
 \label{eq_738132.537765}        & = & H(t) + \left< \log\left( p(t|x)\right) \right>_{p(t,x)}
\end{IEEEeqnarray} 
where $\left< g(x) \right >_{p(x)}$ denotes the expectation of a function of random variable $g(x)$ w.r.t. probability density function $p(x)$; $H(t)$ and $H(t|x)$ are marginal and conditional entropies respectively. Consider the conditional probability of $t$ which is given as  
 \begin{IEEEeqnarray}{rCl}
p(t|x) & = &  \int \dd \theta \dd \gamma \: p( \theta , \gamma, t | x)
 \end{IEEEeqnarray}  
where $\theta$ is a set defined as in~(\ref{eq_738132.476753}). Let $q(\theta , \gamma)$ be an arbitrary distribution. The log conditional probability of $t$ can be expressed as 
 \begin{IEEEeqnarray}{rCl}
  \log(p(t|x)) & = &  \int \dd \theta \dd \gamma \: q(\theta , \gamma) \log\left( p(t|x) \right) \\
   & = &  \int \dd \theta \dd \gamma \: q(\theta , \gamma) \log\left( \frac{p( \theta, \gamma, t | x)}{p( \theta , \gamma | t,x )}   \right)\\
\label{eq_738132.531973} & = &  \int \dd \theta \dd \gamma \: q(\theta , \gamma) \log\left( p( \theta, \gamma, t | x) / q(\theta , \gamma) \right) +  \int \dd \theta \dd \gamma \: q(\theta , \gamma) \log\left( \frac{q(\theta , \gamma)}{p( \theta , \gamma | t,x )}  \right). 
 \end{IEEEeqnarray}    
Define
 \begin{IEEEeqnarray}{rCl}
 \mathcal{L}(q(\theta , \gamma),t,x) & := & \int \dd \theta \dd \gamma \: q(\theta , \gamma) \log\left( \frac{p( \theta, \gamma, t | x)}{q(\theta , \gamma)}   \right)    \IEEEeqnarraynumspace
 \end{IEEEeqnarray}     
to express (\ref{eq_738132.531973}) as
 \begin{IEEEeqnarray}{rCl}
   \log(p(t|x)) & = &  \mathcal{L}(q(\theta , \gamma),t,x)  + \mathrm{KL}(q(\theta , \gamma)\| p(\theta, \gamma | t, x)) \IEEEeqnarraynumspace
 \end{IEEEeqnarray}  
where $\mathrm{KL}$ is Kullback-Leibler divergence of $p(\theta, \gamma | t,x)$ from $q(\theta , \gamma)$. Using (\ref{eq_738132.537765}),
 \begin{IEEEeqnarray}{rCl}
 \label{eq_738132.543734}  I(t;x)& = & H(t) + \left<  \mathcal{L}(q(\theta , \gamma),t,x) \right>_{p(t,x)} + \left< \mathrm{KL}(q(\theta , \gamma)\| p(\theta, \gamma | t, x)) \right>_{p(t,x)}.
 \end{IEEEeqnarray}  
 That is,
  \begin{IEEEeqnarray}{rCl}
 \label{eq_738542.3802}   IL_{f_{t\rightarrow x}} & = & \left<  \mathcal{L}(q(\theta , \gamma),t,x) \right>_{p(t,x)} + \left< \mathrm{KL}(q(\theta , \gamma)\| p(\theta, \gamma | t, x)) \right>_{p(t,x)}.
 \end{IEEEeqnarray}  
Since Kullback-Leibler divergence is always non-zero, it follows from (\ref{eq_738542.3802}) that $\left<  \mathcal{L} \right>_{p(t,x)}$ provides a lower bound on $IL_{f_{t\rightarrow x}}$ i.e.  
  \begin{IEEEeqnarray}{rCl}
\label{eq_738132.547510}IL_{f_{t\rightarrow x}} & \geq &  \left<  \mathcal{L}(q(\theta , \gamma),t,x) \right>_{p(t,x)}.
 \end{IEEEeqnarray} 
Our approach to approximate $IL_{f_{t\rightarrow x}}$ is to maximize its lower bound with respect to variational distribution $q(\theta,\gamma)$. That is, we seek to solve
 \begin{IEEEeqnarray}{rCl} 
\label{eq_738542.4684}\widehat{IL}_{f_{t\rightarrow x}} & = & \max_{q(\theta , \gamma)}\;   \left<  \mathcal{L}(q(\theta , \gamma),t,x) \right>_{p(t,x)}. \end{IEEEeqnarray}   
\begin{result}[Analytical Expression for Information-Leakage]\label{result_738542.4169}
Given the model~(\ref{eq_738132.598922})-(\ref{eq_738132.599647}), $\widehat{IL}_{f_{t\rightarrow x}}$ is given as
\begin{IEEEeqnarray}{rCl}
\nonumber \widehat{IL}_{f_{t\rightarrow x}} &=& - 0.5 q \log(2\pi) + 0.5 q \left\{ \digamma(\bar{a}_{\gamma}) - \log(\bar{b}_{\gamma}) \right\} \\
\nonumber && {-}\: \frac{\bar{a}_{\gamma}}{2\bar{b}_{\gamma}} \sum_{k=1}^q \left<  |t_k - G(x)   \bar{\mathrm{m}}_k |^2 \right >_{p(t,x)}  -  \frac{\bar{a}_{\gamma}}{2\bar{b}_{\gamma}}   \sum_{k=1}^q \left<  Tr\left( (  \bar{\Lambda}_k)^{-1}   (G(x))^T G(x) \right) \right >_{p(x)} \\
\nonumber     & & {-}\: \frac{1}{2}\sum_{k=1}^q \left \{ ( \hat{\mathrm{m}}_k -  \bar{\mathrm{m}}_k)^T \hat{\Lambda}_k  ( \hat{\mathrm{m}}_k -  \bar{\mathrm{m}}_k)   + Tr\left( \hat{ \Lambda}_k (\bar{\Lambda}_k)^{-1}  \right) -     \log \left(\frac{|(\bar{\Lambda}_k)^{-1}|}{| (\hat{\Lambda}_k)^{-1}|}   \right)  \right\}  + \frac{qM}{2}\\
 \label{eq_738133.352191} && {-}\:   \hat{a}_{\gamma} \log \left(\bar{b}_{\gamma} / \hat{b}_{\gamma}  \right) + \log \left( \Gamma(\bar{a}_{\gamma}) / \Gamma(\hat{a}_{\gamma})  \right)  - (\bar{a}_{\gamma} - \hat{a}_{\gamma}) \Psi(\bar{a}_{\gamma}) + (\bar{b}_{\gamma} - \hat{b}_{\gamma}) \left( \bar{a}_{\gamma} / \bar{b}_{\gamma} \right).  
\end{IEEEeqnarray} 
Here, $\digamma(\cdot)$ is the digamma function and the parameters $(\bar{\Lambda}_k,\bar{\mathrm{m}}_k,\bar{a}_{\gamma},\bar{b}_{\gamma})$ satisfy the following:  
\begin{IEEEeqnarray}{rCl} 
\label{eq_738133.353650} \bar{\Lambda}_k & = & \hat{\Lambda}_k + \left(\bar{a}_{\gamma} /  \bar{b}_{\gamma} \right) \left< (G(x))^T G(x)\right>_{p(x)} \\
\label{eq_738133.353865}  \bar{\mathrm{m}}_k & = & ( \bar{\Lambda}_k)^{-1}\left( \hat{\Lambda}_k \hat{\mathrm{m}}_k +  \frac{\bar{a}_{\gamma}}{\bar{b}_{\gamma}}     \left< (G(x) )^T t_k \right>_{p(t,x)} \right) \\
\label{738133.354037} \bar{a}_{\gamma} & = & \hat{a}_{\gamma} + 0.5q \\
\label{738133.354212}  \bar{b}_{\gamma} & = & \hat{b}_{\gamma} + \frac{1}{2} \sum_{k=1}^q \left<  |t_k - G(x)   \bar{\mathrm{m}}_k |^2 \right >_{p(t,x)}   + \frac{1}{2} \sum_{k=1}^q \left<  Tr\left( (  \bar{\Lambda}_k)^{-1}   (G(x))^T G(x) \right) \right >_{p(x)}. 
    \end{IEEEeqnarray}     
\end{result}
\begin{proof}[Proof of Result~\ref{result_738542.4169}]
Consider     
\begin{IEEEeqnarray}{rCl} 
\label{eq_738132.554496}  \mathcal{L}(q(\theta , \gamma),t,x) &  =  & \left< \log(p(t  | \theta, \gamma, x)) \right>_{q(\theta , \gamma)}  +  \left< \log\left( p(\theta , \gamma )/ q(\theta, \gamma) \right) \right>_{q(\theta , \gamma)}.
  \end{IEEEeqnarray}   
It follows from (\ref{eq_738132.598922}) and (\ref{eq_738132.599466}) that
\begin{IEEEeqnarray}{rCl}
 \log(p(t_k  |  \theta_k , \gamma, x))
& = & -0.5\log(2\pi) + 0.5\log(\gamma)  - 0.5 \gamma |t_k - G(x) \theta_k  |^2.
 \end{IEEEeqnarray}  
Since $t = \left[\begin{IEEEeqnarraybox*}[][c]{,c/c/c,}  t_1 & \cdots &t_q
 \end{IEEEeqnarraybox*} \right]^T$, we have
\begin{IEEEeqnarray}{rCl}
\label{eq_738132.613731} \log(p(t | \theta , \gamma, x))& = & -0.5 q \log(2\pi) + 0.5 q \log(\gamma)  - 0.5 \gamma \sum_{k=1}^q |t_k - G(x) \theta_k  |^2.
\end{IEEEeqnarray} 
Using~(\ref{eq_738132.613731}) and (\ref{eq_738132.721205}-\ref{eq_738132.721515}) in (\ref{eq_738132.554496}), we have 
  \begin{IEEEeqnarray}{rCl} 
\nonumber \mathcal{L}(q(\theta , \gamma),t,x) &  =  & -\frac{q}{2} \log(2\pi) + \frac{q}{2} \left< \log(\gamma) \right>_{q(\gamma)}   - \frac{ \left< \gamma \right>_{q(\gamma)}}{2}  \sum_{k=1}^q \left<  |t_k - G(x) \theta_k  |^2\right>_{q(\theta_k)}   \\
&& {+}\: \sum_{k=1}^q \left< \log\left( \frac{p(\theta_k  ; \hat{\mathrm{m}}_k, \hat{\Lambda}_k)}{q(\theta_k )}   \right) \right>_{q(\theta_k )}   +    \left< \log\left( \frac{p(\gamma  ;  a_{\gamma}, b_{\gamma})}{q( \gamma)}   \right) \right>_{q( \gamma)}. 
   \end{IEEEeqnarray} 
Thus,  
  \begin{IEEEeqnarray}{rCl} 
\nonumber \left<  \mathcal{L}(q(\theta , \gamma),t,x) \right>_{p(t,x)} & = &   -\frac{q}{2}\log(2\pi) +  \frac{q}{2} \left< \log(\gamma) \right>_{q(\gamma)}  -  \frac{  \left< \gamma \right>_{q(\gamma)} }{2} \sum_{k=1}^q \left< |t_k|^2 \right>_{p(t)} \\
\nonumber  && {-}\: \frac{\left< \gamma \right>_{q(\gamma)}}{2}  \sum_{k=1}^q \left< (\theta_k)^T \left< (G(x))^T G(x)\right>_{p(x)} \theta_k \right >_{q(\theta_k)} + \left< \gamma \right>_{q(\gamma)} \sum_{k=1}^q \left< (\theta_k)^T \left< (G(x) )^T t_k \right>_{p(t,x)}  \right >_{q(\theta_k)} \\
 && {+}\: \sum_{k=1}^q \left< \log\left( \frac{p(\theta_k  ; \hat{\mathrm{m}}_k, \hat{\Lambda}_k)}{q(\theta_k )}  \right) \right>_{q(\theta_k )}   +  \left< \log\left( \frac{p(\gamma  ;  a_{\gamma}, b_{\gamma}) }{q( \gamma)}  \right) \right>_{q( \gamma)}.  
\end{IEEEeqnarray} 
Now, $\left<  \mathcal{L}(q(\theta , \gamma),t,x) \right>_{p(t,x)}$ can be maximized w.r.t. $q(\theta_k)$ and $q(\gamma)$ using variational optimization. It can be seen that optimal distributions maximizing $\left<  \mathcal{L}(q(\theta , \gamma),t,x) \right>_{p(t,x)}$ are given as     
 \begin{IEEEeqnarray}{rCl} 
 q^*(\theta_k ) & = & \frac{1}{\sqrt{(2\pi)^M | (\bar{\Lambda}_k)^{-1}|}} \exp\left(-0.5(\theta_k  - \bar{\mathrm{m}}_k)^T \bar{\Lambda}_k (\theta_k  - \bar{\mathrm{m}}_k) \right)  \\
  q^*(\gamma ) & = & \left((\bar{b}_{\gamma})^{\bar{a}_{\gamma}} / \Gamma(\bar{a}_{\gamma}) \right) (\gamma)^{\bar{a}_{\gamma}-1} \exp(-\bar{b}_{\gamma} \gamma)
    \end{IEEEeqnarray} 
where the parameters $(\bar{\Lambda}_k,\bar{\mathrm{m}}_k,\bar{a}_{\gamma},\bar{b}_{\gamma})$ satisfy (\ref{eq_738133.353650})-(\ref{738133.354212}). The maximum attained value of $\left<  \mathcal{L}(q(\theta , \gamma),t,x) \right>_{p(t,x)}$ is given as  
\begin{IEEEeqnarray}{rCl} 
 \nonumber  \max_{q(\theta , \gamma)}\; \left<  \mathcal{L}(q(\theta , \gamma),t,x) \right>_{p(t,x)}    &= & - 0.5q\log(2\pi) + 0.5q\left\{ \digamma(\bar{a}_{\gamma}) - \log(\bar{b}_{\gamma}) \right\}  - \frac{\bar{a}_{\gamma}}{2\bar{b}_{\gamma}} \sum_{k=1}^q \left<  |t_k - G(x)  \bar{\mathrm{m}}_k |^2 \right >_{p(t,x)} \\
\nonumber     & & {-} \: \frac{\bar{a}_{\gamma}}{2\bar{b}_{\gamma}} \sum_{k=1}^q \left<  Tr\left( (  \bar{\Lambda}_k)^{-1}   (G(x))^T G(x) \right) \right >_{p(x)}  - \sum_{k=1}^q \mathrm{KL}(q^*(\theta_k)\|  p(\theta_k ; \hat{\mathrm{m}}_k, \hat{\Lambda}_k)) \\
\nonumber && - \mathrm{KL}(q^*(\gamma)\|  p(\gamma  ;  \hat{a}_{\gamma}, \hat{b}_{\gamma}))
\end{IEEEeqnarray}        
where $\digamma(\cdot)$ is the digamma function. After substituting the maximum value in (\ref{eq_738542.4684}) and calculating Kullback-Leibler divergences, we get (\ref{eq_738133.352191}).   
\end{proof}
\subsection{An Algorithm for Computing Information-Leakage}
Result~\ref{result_738542.4169} forms the basis of developing an algorithm for practically computing information-leakage using available data samples.
\begin{algorithm}
\caption{Estimation of information-leakage, $ IL_{f_{t\rightarrow x}} = I(t;x) - H(t)$, using variational approximation}
\label{algorithm_estimation_mutual_information}
\begin{algorithmic}[1]
\Require  Dataset $\left\{ (x^i \in \mathbb{R}^n,t^i \in \mathbb{R}^q) \; | \; x^i = f_{t\rightarrow x}(t^i),\; i \in \{1,\cdots,N \} \right \}$. 
\State Apply Algorithm~\ref{algorithm_variational_Bayesian_learning} on $\left\{ (x^i ,t^i ) \; | \;  i \in \{1,\cdots,N \} \right \}$ with $M_{max} = \min(\lceil N/2 \rceil,1000)$ to obtain variational membership-mappings Bayesian model $\mathbb{BM}^{x \rightarrow t} =  \{\{\hat{\mathrm{m}}_k,\hat{\Lambda}_k\;|\;k \in \{1,\cdots,q \} \},\hat{a}_{\gamma},\hat{b}_{\gamma} \} $.   
\State Initialise $\bar{a}/\bar{b} = \hat{a}/\hat{b}$.
\Repeat
\State Update $\{\bar{\Lambda}_k,\bar{\mathrm{m}}_k\;|\;k \in \{1,\cdots,q \} \},\bar{a},\bar{b}$ using (\ref{eq_738133.353650})-(\ref{738133.354212}) where expectations $<\cdot>_{p(x)}$ and $<\cdot>_{p(t,x)}$ are approximated via sample-averages.
 \Until{convergence.}
 \State Compute $\widehat{IL}_{f_{t\rightarrow x}}$ using (\ref{eq_738133.352191}) where expectations $<\cdot>_{p(x)}$ and $<\cdot>_{p(t,x)}$ are approximated via sample-averages. 
 \State \Return $\widehat{IL}_{f_{t\rightarrow x}}$ and the model $\mathbb{BM}^{x \rightarrow t}$.
\end{algorithmic} 
\end{algorithm}  
\begin{example}[Verification of Information-Leakage Estimation Algorithm]
To demonstrate the effectiveness of Algorithm~\ref{algorithm_estimation_mutual_information} for estimating information-leakage, a scenario is generated where $t \in \mathbb{R}^{10}$ and $x \in \mathbb{R}^{10}$ are Gaussian distributed such that $x = t + \omega$; $t  \sim  \mathcal{N}(0,5I_{10})$; $\omega \sim \mathcal{N}(0,\sigma I_{10})$ with $\sigma \in [1,15] $. Since the data distributions in this scenario are known, the information-leakage can be theoretically calculated and is given as  
 \begin{IEEEeqnarray}{rCl} 
\nonumber  IL_{f_{t\rightarrow x}} & = & 5  \log(1 + 5/\sigma) - 0.5\log\left(|(2\pi \mathrm{e} 5I_{10})|\right).
 \end{IEEEeqnarray}  
 For a given value of $\sigma$, 1000 samples of $t$ and $x$ were simulated and Algorithm~\ref{algorithm_estimation_mutual_information} was applied for estimating information-leakage. The experiments were carried out at different values of $\sigma$ ranging from $1$ to $15$.
 \begin{figure}[!h]
\centering
\includegraphics[width = 0.7\textwidth]{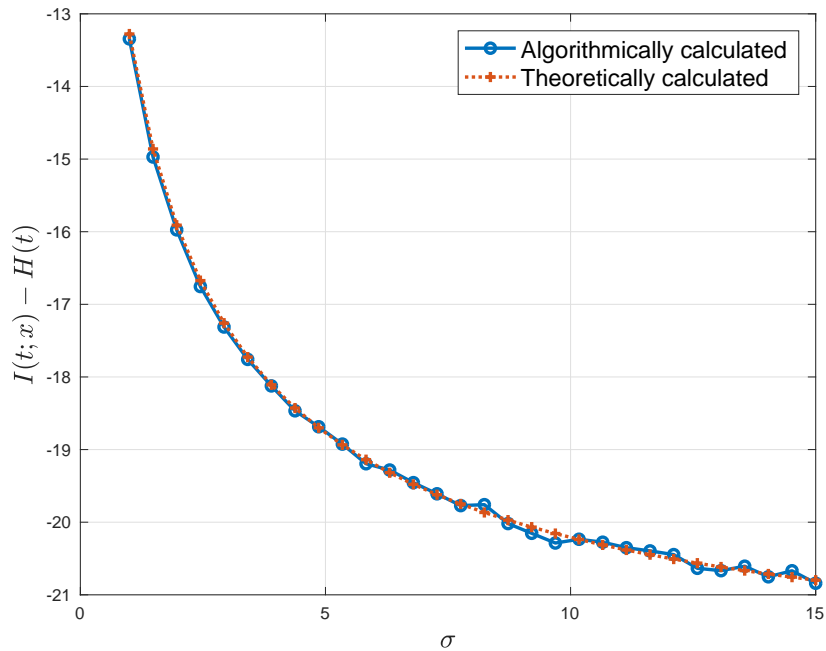}
\caption{A comparison of the estimated information-leakage values with the theoretically calculated values.}
\label{fig_MI_estimation}
\end{figure}    
Fig.~\ref{fig_MI_estimation} compares the plots of estimated and theoretically calculated values of information-leakage against $\sigma$. A close agreement between the two plots in Fig.~\ref{fig_MI_estimation} verifies the effectiveness of Algorithm~\ref{algorithm_estimation_mutual_information} in estimating information-leakage without knowing the data distributions.  
\end{example}
\section{Information Theoretic Measures for Privacy-Leakage, Interpretability, and Transferability}\label{sec_measures}  
\subsection{Definitions}
To define formally the information theoretic measures for privacy-leakage, interpretability, and transferability; a few variables and mappings are introduced in Table~\ref{table_notations}. Definitions~\ref{def_privacy_leakage}, \ref{def_interpretability}, and \ref{def_transferability} provide the mathematical definitions of the information theoretic measures.
\begin{table}
\caption{Introduced variables and mappings.}
\label{table_notations} \centering
\begin{tabular}{cc}
\toprule
\bfseries symbol/mapping & \bfseries definition/meaning \\ \midrule 
$x_{sr} \in \mathbb{R}^{n_{sr}}$ & $\begin{array}{c} \mbox{vector representing private/sensitive variables associated to source domain} \end{array}$    \\ \midrule
$y_{sr} \in \mathbb{R}^{p_{sr}}$ & $\begin{array}{c} \mbox{source domain data vector}  \end{array}$      \\  \midrule
 $t_{sr} \in \mathbb{R}^q$ & $\begin{array}{c} \mbox{vector representing the set of interpretable parameters associated to} \\ \mbox{non-interpretable data vector $y_{sr}$}  \end{array}$ \\ \midrule 
 $y_{sr}^+  \in \mathbb{R}^{p_{sr}}$ & $\begin{array}{c} \mbox{noise added data vector (that is either publicly released or used for the training of} \\ \mbox{source model) obtained from $y_{sr}$ via Algorithm~\ref{algorithm_differential_private_approximation}} \end{array}$     \\ \midrule
 $f_{x_{sr} \rightarrow y_{sr}^+}:\mathbb{R}^{n_{sr}} \rightarrow  \mathbb{R}^{p_{sr}}$ & $\begin{array}{c} \mbox{mapping from private variables to noise added data vector, i.e., $y_{sr}^+ = f_{x_{sr} \rightarrow y_{sr}^+}(x_{sr})$} \end{array}$  \\ \midrule
  $f_{t_{sr} \rightarrow y_{sr}^+}:\mathbb{R}^{q} \rightarrow  \mathbb{R}^{p_{sr}}$ & $\begin{array}{c} \mbox{mapping from interpretable parameters to noise added data vector, i.e., $y_{sr}^+ = f_{t_{sr} \rightarrow y_{sr}^+}(t_{sr})$} \end{array}$  \\ \midrule
 $ \{ \mathcal{P}_c^{+sr} \}_{c= 1}^C$ & $\begin{array}{c} \mbox{differentially private source domain autoencoders, representing data features of} \\ \mbox{each of $C$ classes, obtained via Algorithm~\ref{algorithm_private_classification}}  \end{array}$  \\ \midrule
  $y_{tg} \in \mathbb{R}^{p_{tg}}$ & target domain data vector \\  \midrule
   $y_{tg \rightarrow sr}  \in \mathbb{R}^{p_{sr}}$ & $\begin{array}{c} \mbox{representation of target domain data vector $y_{tg}$ in source domain via transformation (\ref{eq_1643012238})} \end{array}$ \\ \midrule
   $ \{ \mathcal{P}_c^{tg} \}_{c= 1}^C$ & $\begin{array}{c} \mbox{target domain autoencoders, representing data features of} \\ \mbox{each of $C$ classes, obtained via Algorithm~\ref{algorithm_private_classification}}  \end{array}$  \\ \midrule
 $f_{y_{tg} \rightarrow c} : \mathbb{R}^{p_{tg}} \rightarrow \{1,\cdots,C \}$ & $\begin{array}{c} \mbox{mapping assigning class-label to target domain data vector $y_{tg}$ via (\ref{eq_predicted_target_label_multitask}), i.e.,} \\ \mbox{$f_{y_{tg} \rightarrow c}(y_{tg}) =  \hat{c}\left(y_{tg \rightarrow sr}(y_{tg});\{ \mathcal{P}_c^{tg} \}_{c=1}^C, \{ \mathcal{P}_c^{+sr} \}_{c=1}^C,  \mathbb{M}^{sr\rightarrow tg}\right)$} \end{array}$ \\  \midrule
$\hat{y}_{tg}^{sr}  \in \mathbb{R}^{p_{sr}}$ & $\begin{array}{c} \mbox{transformation of $y_{tg}$ to source domain and filtering through the autoencoder} \\ \mbox{that represents the source domain feature vectors of the same class as that of $y_{tg}$, i.e.,} \\ \hat{y}_{tg}^{sr} =  \widehat{\mathcal{WD}}\left(y_{tg \rightarrow sr}(y_{tg});\mathcal{P}_{f_{y_{tg} \rightarrow c}(y_{tg}) }^{+sr}\right) \end{array}$ \\ \midrule
$\hat{y}_{tg}^{tg}  \in \mathbb{R}^{p_{sr}}$ & $\begin{array}{c} \mbox{transformation of $y_{tg}$ to source domain and filtering through the autoencoder} \\ \mbox{that represents the target domain feature vectors of the same class as that of $y_{tg}$, i.e.,} \\ \hat{y}_{tg}^{tg} =  \widehat{\mathcal{WD}}\left(y_{tg \rightarrow sr}(y_{tg});\mathcal{P}_{f_{y_{tg} \rightarrow c}(y_{tg}) }^{tg}\right) \end{array}$ \\ \midrule
$f_{\hat{y}_{tg}^{sr} \rightarrow \hat{y}_{tg}^{tg}}: \mathbb{R}^{p_{sr}} \rightarrow \mathbb{R}^{p_{sr}} $ & $\begin{array}{c} \mbox{mapping from source domain feature vector $ \hat{y}_{tg}^{sr} $ to target domain feature vector $\hat{y}_{tg}^{tg}$, i.e.,} \\ \hat{y}_{tg}^{tg}  =   f_{\hat{y}_{tg}^{sr} \rightarrow \hat{y}_{tg}^{tg}} \left( \hat{y}_{tg}^{sr}  \right)\end{array}$ \\
 \bottomrule
\end{tabular}
\end{table}    
\begin{definition}[Privacy-Leakage]\label{def_privacy_leakage}
Privacy-leakage (by the mapping from private variables to noise added data vector) is a measure of the amount of information about private/sensitive variable $x_{sr}$ leaked by the mapping $f_{x_{sr} \rightarrow y_{sr}^+}$ and is defined as   
 \begin{IEEEeqnarray}{rCl} 
 IL_{f_{x_{sr} \rightarrow y_{sr}^+}} & := & I\left(x_{sr};f_{x_{sr} \rightarrow y_{sr}^+}(x_{sr})\right) - H(x_{sr}) \\
 & = &  I\left(x_{sr};y_{sr}^+\right) - H(x_{sr}).
 \end{IEEEeqnarray}  
\end{definition} 
\begin{definition}[Interpretability-Measure]\label{def_interpretability}
Interpretability (of noise added data vector) is measured as the amount of information about interpretable parameters $t_{sr}$ leaked by the mapping $f_{t_{sr} \rightarrow y_{sr}^+}$ and is defined as 
 \begin{IEEEeqnarray}{rCl} 
 IL_{f_{t_{sr} \rightarrow y_{sr}^+}} & := & I\left(t_{sr};f_{t_{sr} \rightarrow y_{sr}^+}(t_{sr})\right) - H(t_{sr}) \\
 & = &  I\left(t_{sr};y_{sr}^+\right) - H(t_{sr}).
 \end{IEEEeqnarray}  
\end{definition}
\begin{definition}[Transferability-Measure]\label{def_transferability}
Transferability (from source domain data representation learning models (i.e. $ \mathcal{P}_1^{+sr},\cdots,\mathcal{P}_C^{+sr}$) to the target domain data representation learning models (i.e. $ \mathcal{P}_1^{tg},\cdots,\mathcal{P}_C^{tg}$)) is measured as the amount of information about source domain feature vector $ \hat{y}_{tg}^{sr} $ leaked by the mapping $f_{\hat{y}_{tg}^{sr} \rightarrow \hat{y}_{tg}^{tg}}$ and is defined as   
 \begin{IEEEeqnarray}{rCl} 
 IL_{f_{\hat{y}_{tg}^{sr} \rightarrow \hat{y}_{tg}^{tg}}} & := & I\left( \hat{y}_{tg}^{sr} ;f_{\hat{y}_{tg}^{sr} \rightarrow \hat{y}_{tg}^{tg}}( \hat{y}_{tg}^{sr})\right) - H( \hat{y}_{tg}^{sr} ) \\
 & = & I\left( \hat{y}_{tg}^{sr} ;\hat{y}_{tg}^{tg} \right) - H( \hat{y}_{tg}^{sr} ).
 \end{IEEEeqnarray}  
 Here, $\hat{y}_{tg}^{tg}$ represents the target domain feature vector and $f_{\hat{y}_{tg}^{sr} \rightarrow \hat{y}_{tg}^{tg}}: \mathbb{R}^{p_{sr}} \rightarrow \mathbb{R}^{p_{sr}} $ is the mapping from source domain feature vector $ \hat{y}_{tg}^{sr} $ to target domain feature vector $\hat{y}_{tg}^{tg}$. 
  \end{definition}

Since the defined measures are in the form of information-leakages, Algorithm~\ref{algorithm_estimation_mutual_information} could be directly applied for practically computing the measures provided the availability of data samples.  
\subsection{A Unified Approach to Privacy-Preserving Interpretable and Transferable Learning}
The presented theory allows to develop an algorithm that implements privacy-preserving interpretable and transferable learning methodology in a unified manner. 
\begin{algorithm}
\caption{Algorithm for privacy-preserving interpretable and transferable learning}
\label{algorithm_TAI}
\begin{algorithmic}[1]
\Require The labelled source dataset: $\mathbf{Y}^{sr} = \{\mathbf{Y}^{sr}_c \}_{c=1}^C$ (where $\mathbf{Y}^{sr}_c = \{ y^{i,c}_{sr} \in \mathbb{R}^{p_{sr}} \; | \; i \in \{1,\cdots,N^{sr}_c \} \}$ represents $c-$th labelled samples); the set of private data: $\mathbf{X}^{sr} = \{\mathbf{X}^{sr}_c \}_{c=1}^C$ (where $\mathbf{X}^{sr}_c = \{x_{sr} \in \mathbb{R}^{n_{sr}} \; | \; x_{sr} = f^{-1}_{x_{sr} \rightarrow y_{sr}}(y_{sr}),\; y_{sr} \in  \mathbf{Y}^{sr}_c   \}$); the set of interpretable parameters: $\mathbf{T}^{sr} = \{\mathbf{T}^{sr}_c \}_{c=1}^C$ (where $\mathbf{T}^{sr}_c = \{t_{sr} \in \mathbb{R}^{q} \; | \; t_{sr} = f^{-1}_{t_{sr} \rightarrow y_{sr}}(y_{sr}),\; y_{sr} \in \mathbf{Y}^{sr}_c   \}$); the set of a few labelled target samples: $\{\mathbf{Y}^{tg}_c \}_{c=1}^C$ (where $\mathbf{Y}^{tg}_c = \{ y^{i,c}_{tg} \in \mathbb{R}^{p_{tg}} \; | \; i \in \{1,\cdots,N^{tg}_c \} \}$ is the set of $c-$th labelled target samples); the set of unlabelled target samples: $\mathbf{Y}^{tg}_* = \{ y^{i,*}_{tg} \in \mathbb{R}^{p_{tg}} \; | \; i \in \{1,\cdots,N^{tg}_* \} \}$; and the differential privacy parameters: $d  \in \mathbb{R}_{+}$,  $\epsilon  \in \mathbb{R}_{+}$, $\delta \in (0,1)$.
\State A differentially private approximation of source dataset, $\mathbf{Y}^{+sr} = \{\mathbf{Y}^{+sr}_c \}_{c=1}^C$, is obtained using Algorithm~\ref{algorithm_differential_private_approximation} on $\mathbf{Y}^{sr}$.  
\State Differentially private source domain classifier, $ \{ \mathcal{P}_c^{+sr}  \}_{c=1}^C$, is built using Algorithm~\ref{algorithm_private_classification} on $\mathbf{Y}^{+sr}$ taking subspace dimension as equal to $\min(20,p_{sr})$ (where $p_{sr}$ is the dimension of source data samples), ratio $r_{max}$ as equal to 0.5, and number of layers as equal to 5. 
\State Taking subspace dimension $n_{st} = \min(\lceil p_{sr}/2 \rceil,p_{tg})$, the source domain transformation-matrix $V^{+sr} \in \mathbb{R}^{n_{st}  \times p_{sr}}$ is defined as with its $i-$th row equal to transpose of eigenvector corresponding to $i-$th largest eigenvalue of sample covariance matrix computed on differentially private approximated source samples. The target domain transformation-matrix $V^{tg} \in \mathbb{R}^{n_{st} \times p_{tg}}$ is defined as with its $i-$th row equal to transpose of eigenvector corresponding to $i-$th largest eigenvalue of sample covariance matrix computed on target samples. 
\State For the case of heterogenous source and target domains, the subspace alignment approach is used to transform target samples via (\ref{eq_738522.553647}) and (\ref{eq_738522.553880}) for defining the sets $\{ \mathbf{Y}^{tg \rightarrow sr}_c\}_{c=1}^C$ and $\mathbf{Y}^{tg \rightarrow sr}_*$.    
\State Initial target domain classifier, $ \{ \mathcal{P}_c^{tg}|_0  \}_{c=1}^C$, is built using Algorithm~\ref{algorithm_classification} on labelled target samples, $\{ \mathbf{Y}^{tg \rightarrow sr}_c\}_{c=1}^C$, taking subspace dimension as equal to $\min(20,\min_{1\leq c \leq C}\{ N^{tg}_c\}-1)$ (where $N^{tg}_c$ is the number of $c-$th class labelled target samples), ratio $r_{max}$ as equal to 1, and number of layers as equal to 1. 
\State The target domain classifier is updated using (\ref{eq_iterative_target_classification}) and (\ref{eq_iterative_target_samples_class_c}) till 4 iterations taking the monotonically non-decreasing subspace dimension $n$ sequence as $\{\min(5,p_{sr}),\min(10,p_{sr}),\min(15,p_{sr}),\min(20,p_{sr})\}$ and $r_{max = 0.5}$. 
\State The mapping from source to target domain is learned by means of a model, $\mathbb{M}^{sr\rightarrow tg}$, defined as in (\ref{eq_738522.588899}).
\State  Compute privacy-leakage, $ IL_{f_{x_{sr} \rightarrow y_{sr}^+}}$, and adversary model, $\mathbb{BM}^{y_{sr}^+ \rightarrow x_{sr}}$, via applying Algorithm~\ref{algorithm_estimation_mutual_information} on $\{ (y^{+}_{sr},x_{sr})  \; | \; y_{sr}^{+} = f_{x_{sr} \rightarrow y_{sr}^{+}}(x_{sr}),\; x_{sr} \in \mathbf{X}^{sr}, \; y^{+}_{sr} \in   \mathbf{Y}^{+sr}  \}$.  
\State  Compute interpretability-measure, $ IL_{f_{t_{sr} \rightarrow y_{sr}^+}} $, and interpretability model, $\mathbb{BM}^{y_{sr}^+ \rightarrow t_{sr}}$, via applying Algorithm~\ref{algorithm_estimation_mutual_information} on $\{ (y^{+}_{sr},t_{sr})  \; | \; y_{sr}^{+} = f_{t_{sr} \rightarrow y_{sr}^+}(t_{sr}),\; t_{sr} \in \mathbf{T}^{sr}, \; y^{+}_{sr} \in   \mathbf{Y}^{+sr}  \}$.  
\State Compute transferability-measure, $ IL_{f_{\hat{y}_{tg}^{sr} \rightarrow \hat{y}_{tg}^{tg}}}$, via applying Algorithm~\ref{algorithm_estimation_mutual_information} on $\left\{\left(\hat{y}_{tg}^{tg}(y_{tg}),\hat{y}_{tg}^{sr}(y_{tg})\right) \;  | \; y_{tg} \in \{\mathbf{Y}^{tg}_c \}_{c=1}^C \cup \mathbf{Y}^{tg}_*  \right \}$, where
 \begin{IEEEeqnarray}{rCl} 
 \hat{y}_{tg}^{sr}(y_{tg}) & = &  \widehat{\mathcal{WD}}\left(y_{tg \rightarrow sr}(y_{tg});\mathcal{P}_{f_{y_{tg} \rightarrow c}(y_{tg}) }^{+sr}\right)\\
 \hat{y}_{tg}^{tg}(y_{tg}) & = &  \widehat{\mathcal{WD}}\left(y_{tg \rightarrow sr}(y_{tg});\mathcal{P}_{f_{y_{tg} \rightarrow c}(y_{tg}) }^{tg}\right)\\
 f_{y_{tg} \rightarrow c}(y_{tg}) & = &  \hat{c}\left(y_{tg \rightarrow sr}(y_{tg});\{ \mathcal{P}_c^{tg} \}_{c=1}^C, \{ \mathcal{P}_c^{+sr} \}_{c=1}^C,  \mathbb{M}^{sr\rightarrow tg}\right),
  \end{IEEEeqnarray}   
$y_{tg \rightarrow sr}(y_{tg})$ is defined as in (\ref{eq_1643012238}), and $\hat{c}(\cdot)$ is defined by (\ref{eq_predicted_target_label_multitask}). 
\State \Return \underline{in the source domain:} classifier $ \{ \mathcal{P}_c^{+sr}  \}_{c=1}^C$; privacy-leakage $ IL_{f_{x_{sr} \rightarrow y_{sr}^+}}$ and adversary model $\mathbb{BM}^{y_{sr}^+ \rightarrow x_{sr}}$; interpretability-measure $ IL_{f_{t_{sr} \rightarrow y_{sr}^+}} $ and interpretability model $\mathbb{BM}^{y_{sr}^+ \rightarrow t_{sr}}$.   
\State \Return \underline{in the target domain:} classifier $ \{ \mathcal{P}_c^{tg}  \}_{c=1}^C$.
\State \Return \underline{for transfer and multi-task learning scenario:} classifiers $ \{ \mathcal{P}_c^{+sr}  \}_{c=1}^C$ and $ \{ \mathcal{P}_c^{tg}  \}_{c=1}^C$; source2target model $\mathbb{M}^{sr\rightarrow tg}$; latent subspace transformation-matrices $V^{+sr}$ and $V^{tg}$; transferability-measure $ IL_{f_{\hat{y}_{tg}^{sr} \rightarrow \hat{y}_{tg}^{tg}}}$.          
\end{algorithmic}
\end{algorithm}
Algorithm~\ref{algorithm_TAI} is presented for a systematic implementation of the proposed privacy-preserving interpretable and transferable deep learning methodology. Algorithm~\ref{algorithm_TAI} provides
\begin{enumerate}
\item information theoretic evaluation of privacy-leakage, interpretability, and transferability in a semi-supervised transfer and multi-task learning scenario;
\item the adversary model $\mathbb{BM}^{y_{sr}^+ \rightarrow x_{sr}}$, that can be used to estimate private data and thus to simulate privacy attacks;
\item the interpretability model $\mathbb{BM}^{y_{sr}^+ \rightarrow t_{sr}}$, that can be used to estimate interpretable parameters and thus to provide an interpretation to the non-interpretable data vectors.   
\end{enumerate}       
\section{Experiments}\label{sec_experiments}
Experiments have been carried out to demonstrate the application of the proposed measures (for privacy-leakage, interpretability, and transferability) to privacy-preserving interpretable and transferable learning. The methodology was implemented using MATLAB R2017b and the experiments have been made on an iMac (M1, 2021) machine with 8 GB RAM.     
\subsection{MNIST Dataset}
The MNIST dataset contains $28 \times 28$ sized images divided into training set of 60000 images and test set of 10000 images. The images' pixel values were divided by 255 to normalize the values in the range from $0$ to $1$. The $28 \times 28$ normalized pixel values of each image were flattened to an equivalent $784-$dimensional data vector. 
\paragraph{Interpretable Parameters:}
For MNIST digits dataset, there exist no additional interpretable parameters other than the pixel values. Thus, we defined corresponding to a pixel values vector $y \in [0,1]^{784}$, an interpretable parameter vector $t \in \{0,1\}^{10}$ such that $j-$th element $t_j = 1$, if $j-$th class-label is associated to $y$, otherwise $t_j = 0$. That is, interpretable vector $t$, in our experimental setting, represents the class-label assigned to data vector $y$. 
\paragraph{Private Data:}
Here we assume that pixel values are private, i.e., $x_{sr} = y_{sr}$.   
\paragraph{Semi-Supervised Transfer Learning Scenario:}
A transfer learning scenario was considered in the same setting as in~\cite{conf/iclr/PapernotAEGT17,kumar2022differentially} where 60000 training samples constituted the source dataset; a set of 9000 test samples constituted target dataset, and the classification performance was evaluated on the remaining 1000 test samples. Out of 9000 target samples, only 10 samples per class were labelled and rest 8900 target samples remained as unlabelled.  
\paragraph{Experimental Design:}
Algorithm~\ref{algorithm_TAI} is applied with the differential privacy parameters as $d=1$, $\epsilon \in \{0.1,0.25,0.5,1,2,10\}$, and $\delta = 1\mathrm{e}{-5}$. The experiment involves 6 different privacy-preserving semi-supervised transfer learning scenarios with privacy-loss bound values as $\epsilon = 0.1$, $\epsilon = 0.25$, $\epsilon = 0.5$, $\epsilon = 1$, $\epsilon = 2$, and $\epsilon = 10$. For the computation of privacy-leakage, interpretability-measure, and transferability-measure in Algorithm~\ref{algorithm_TAI}, a subset consisting of 5000 randomly selected samples was considered.  
\paragraph{Results:}
\begin{figure}
\centerline{ \subfigure[privacy-leakage vs. accuracy]{\includegraphics[width=0.5\textwidth]{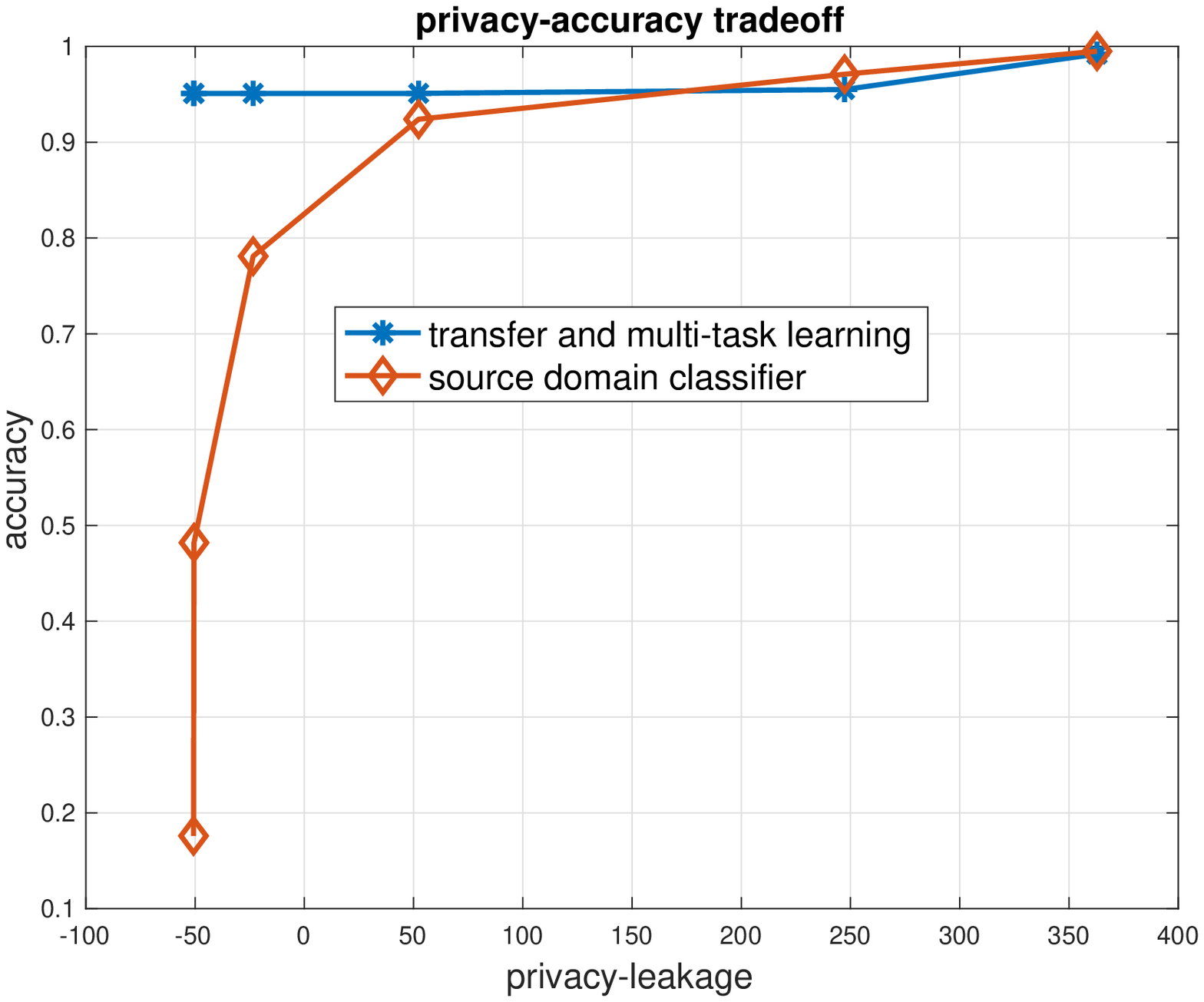}\label{fig-privacy-accuracy}} \hfil  \subfigure[privacy-leakage vs. interpretability-measure]{\includegraphics[width=0.5\textwidth]{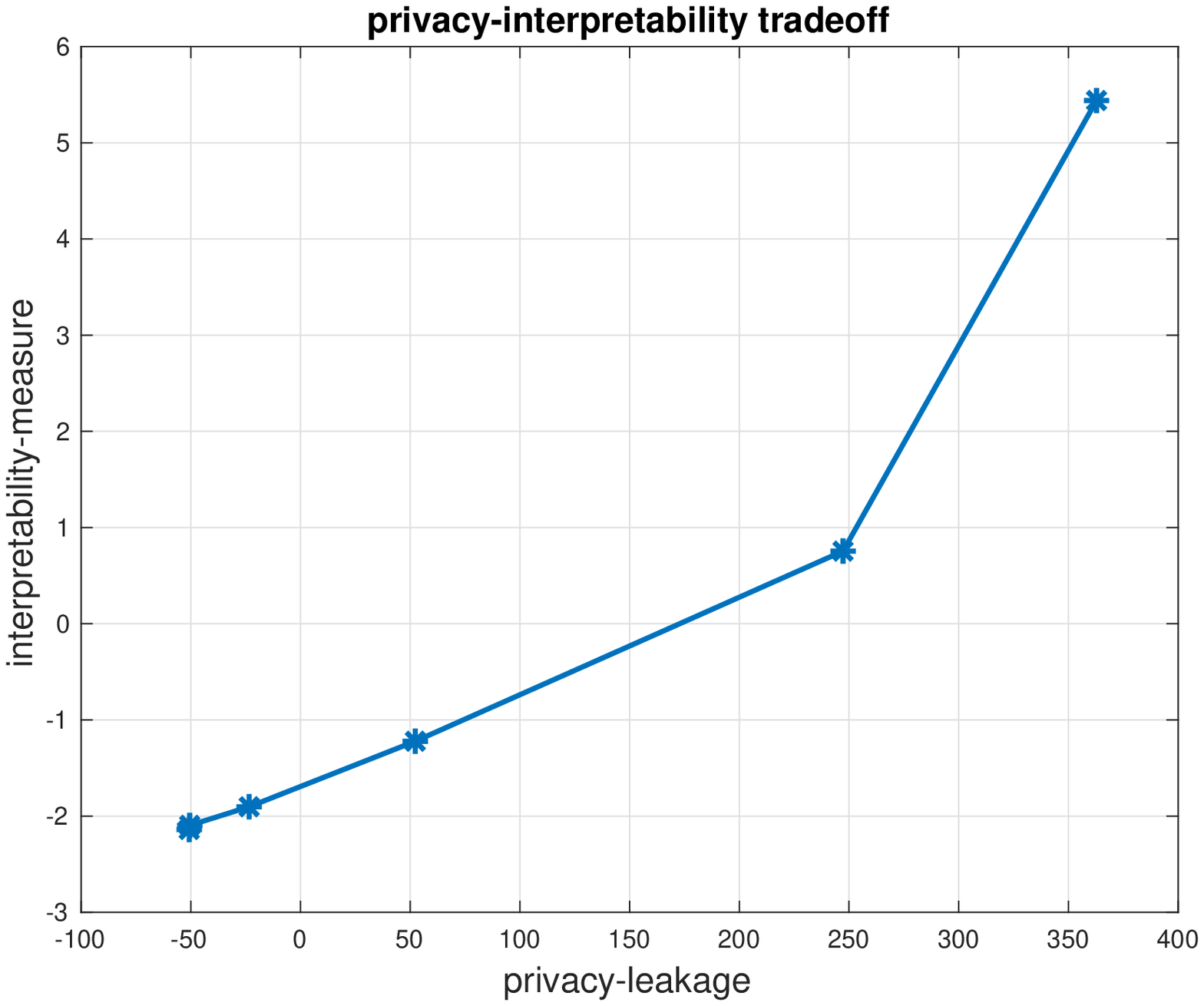}\label{fig-privacy-interpretability}}}
\centerline{ \subfigure[privacy-leakage vs. transferability-measure]{\includegraphics[width=0.5\textwidth]{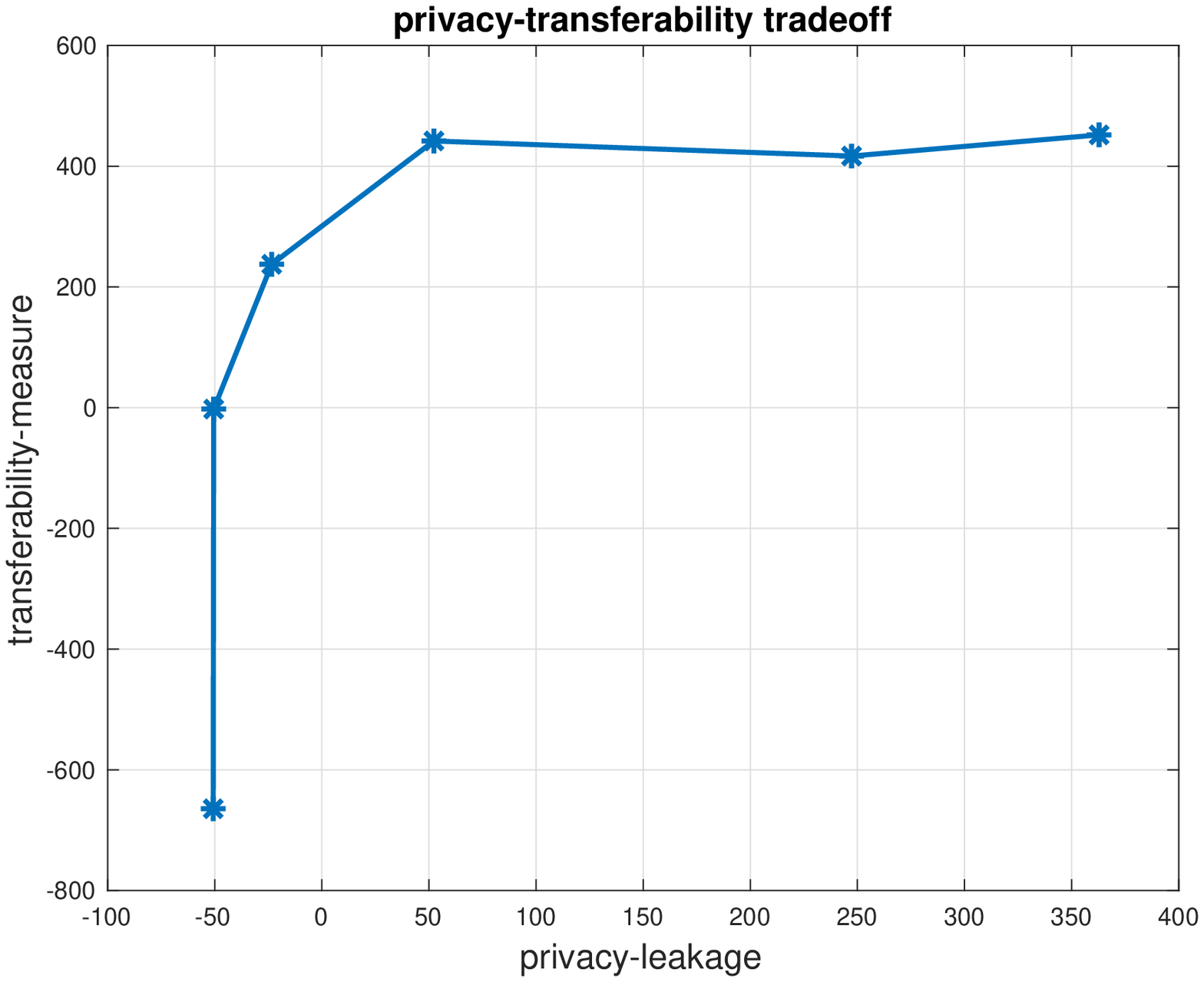}\label{fig-privacy-transferability}} \hfil  \subfigure[interpretability-measure vs. transferability-measure]{\includegraphics[width=0.5\textwidth]{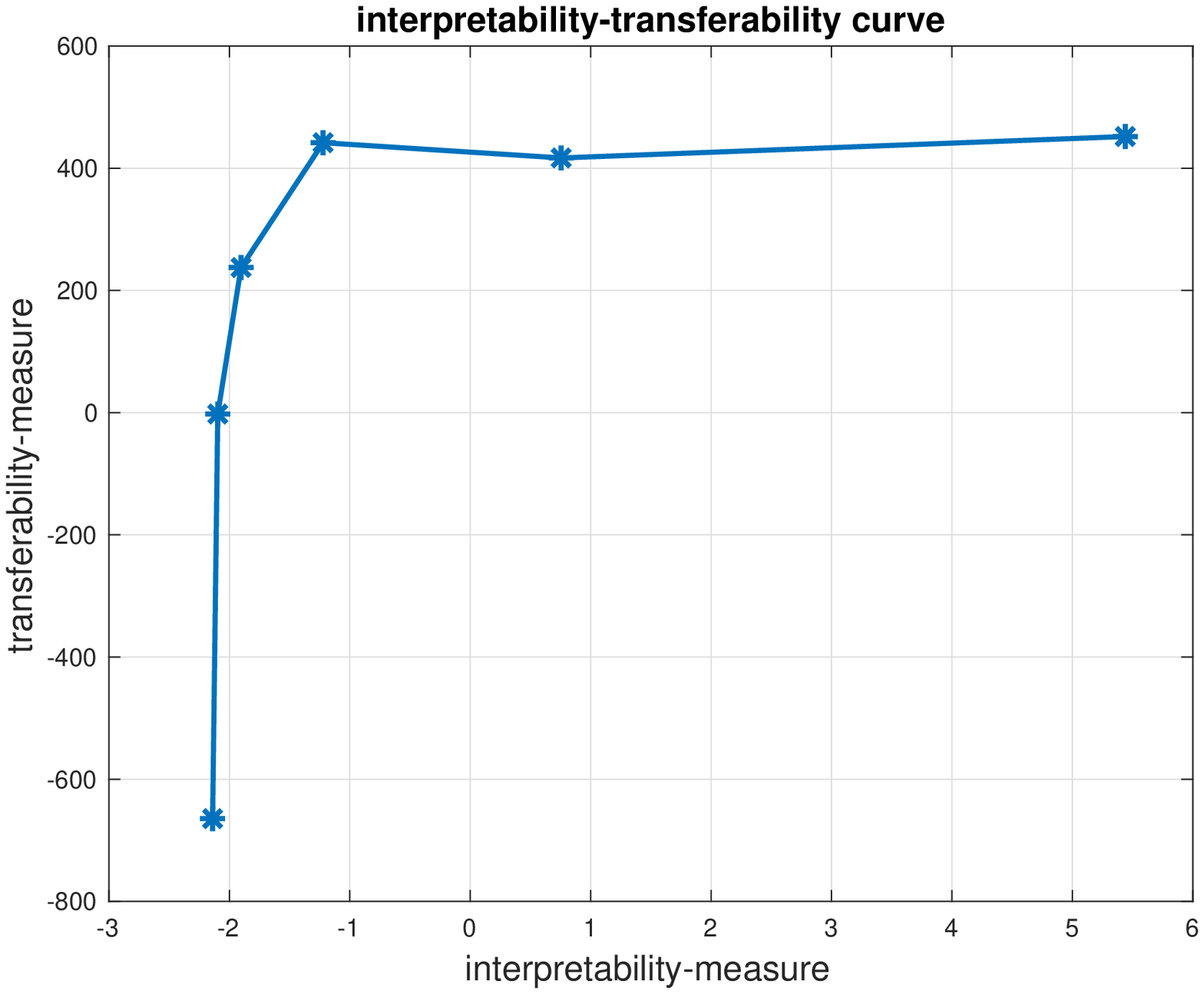}\label{fig-interpretability-transferability}}}
\centerline{ \subfigure[interpretability-measure vs. accuracy]{\includegraphics[width=0.5\textwidth]{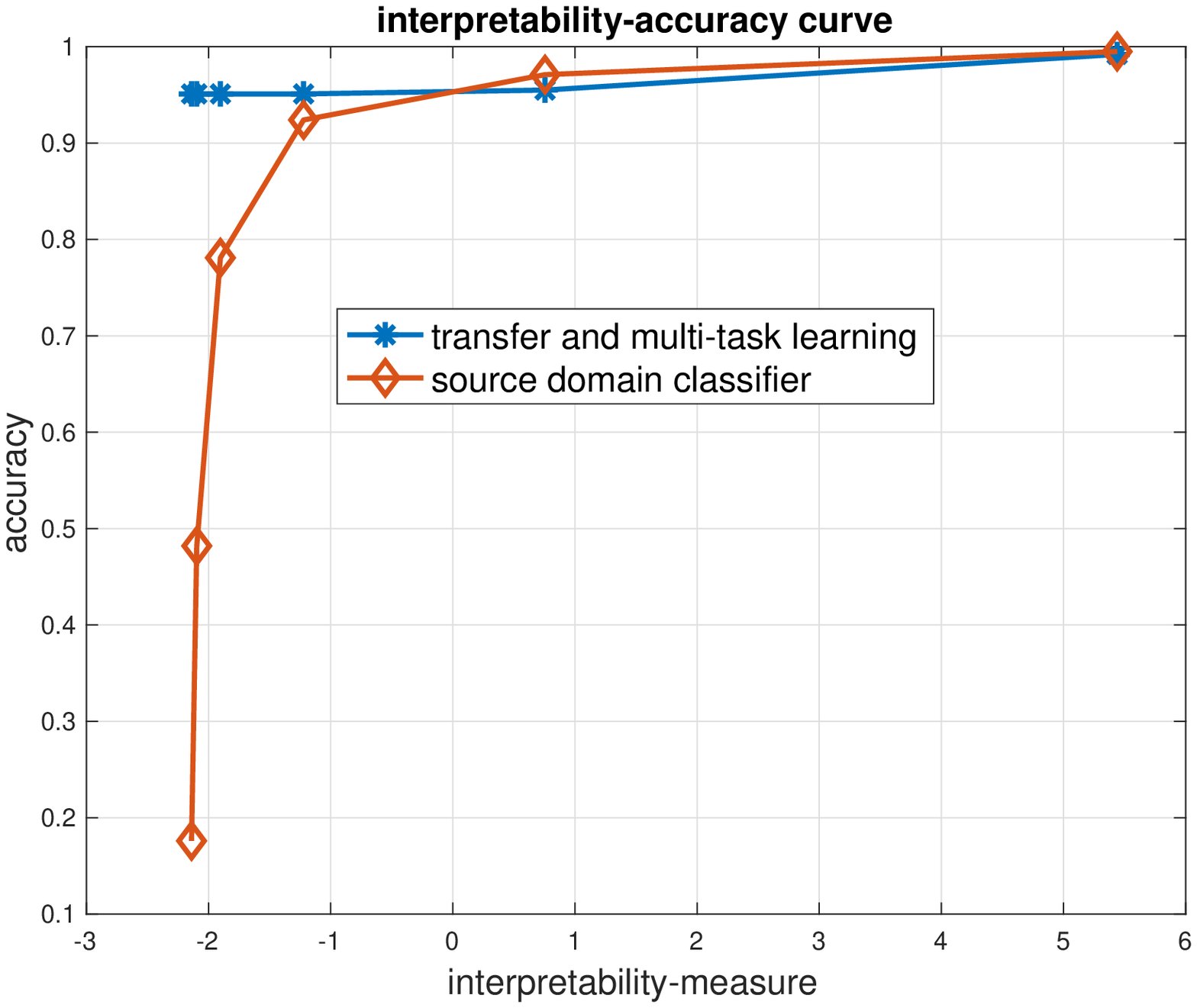}\label{fig-interpretability-accuracy}} \hfil \subfigure[transferability-measure vs. accuracy]{\includegraphics[width=0.5\textwidth]{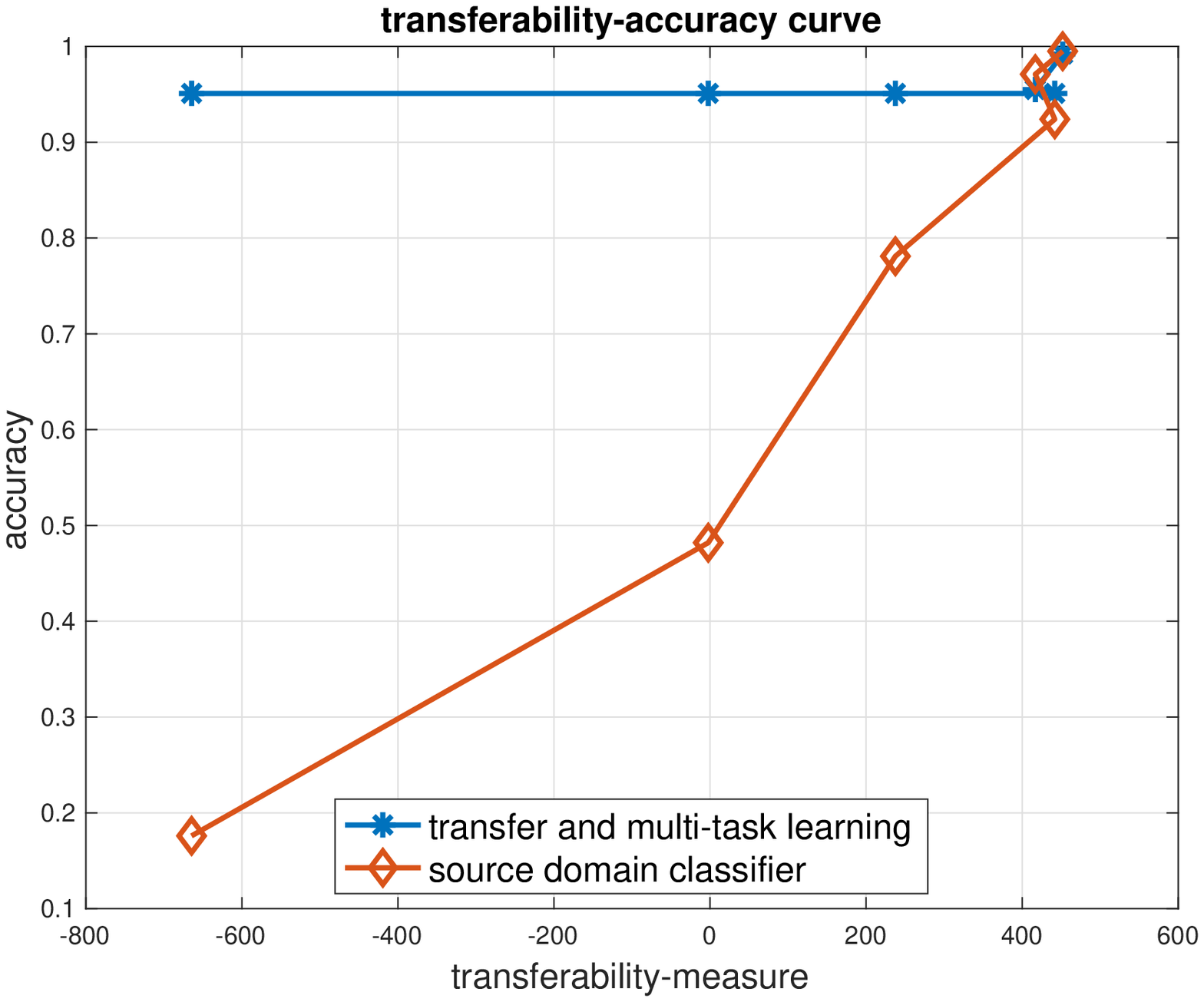}\label{fig-transferability-accuracy}}}
\caption{The plots between privacy-leakage, interpretability-measure, transferability-measure, and accuracy for MNIST dataset.}
\label{fig_TAI_results1}
\end{figure} 
The experimental results have been plotted in Fig.~\ref{fig_TAI_results1}. Fig.~\ref{fig-privacy-accuracy}, Fig.~\ref{fig-privacy-interpretability}, and Fig.~\ref{fig-privacy-transferability} display the privacy-accuracy tradeoff curve, privacy-interpretability tradeoff curve, and privacy-transferability tradeoff curve respectively. As expected and observed in Fig.~\ref{fig-transferability-accuracy}, the transferability-measure is positively correlated with the accuracy of source-domain classifier on target test samples. Since we have defined the interpretable vector associated to a feature vector as representing the class-label, the positive correlations of interpretability-measure with the source domain classifier's accuracy and the transferability-measure are observed in Fig.~\ref{fig-interpretability-accuracy} and Fig.~\ref{fig-interpretability-transferability} respectively. The results also verify the robust performance of Algorithm~\ref{algorithm_TAI} under transfer and multi-task learning scenario, since the classification performance in transfer and multi-task learning scenario, unlike the performance of source domain classifier, is not adversely affected by a reduction in privacy-leakage, interpretability-measure, and transferability-measure as observed in Fig.~\ref{fig-privacy-accuracy}, Fig.~\ref{fig-interpretability-accuracy}, and Fig.~\ref{fig-transferability-accuracy}.               
\begin{table}
\centering
 \caption{Results of experiments on MNIST dataset for evaluating privacy-leakage, interpretability, and transferability.}
 \label{table_TAI_results_MNIST}
  {%
    \begin{tabular}{ccccc}  
    \hline 
    \bfseries Method & \bfseries  $\begin{array}{c} \mbox{privacy-} \\ \mbox{leakage} \end{array}$  & \bfseries $\begin{array}{c} \mbox{interpretability-} \\ \mbox{measure} \end{array}$ & \bfseries $\begin{array}{c} \mbox{transferability-} \\ \mbox{measure} \end{array}$ & \bfseries $\begin{array}{c} \mbox{classification} \\ \mbox{accuracy} \end{array}$   \\  
    \hline 
       $\begin{array}{c} \mbox{minimum privacy-leakage} \\ \mbox{transfer and multi-task learning} \end{array}$ & -50.72  & -2.14 & -664.52 & 0.9510  \\ \hdashline
       $\begin{array}{c} \mbox{minimum privacy-leakage} \\ \mbox{source domain classifier} \end{array}$ & -50.72  & -2.14 & -664.52 & 0.1760 \\ \hline
       $\begin{array}{c} \mbox{maximum interpretability-measure} \\ \mbox{transfer and multi-task learning} \end{array}$ & 362.83  & 5.44 & 451.93 & 0.9920   \\ \hdashline
       $\begin{array}{c} \mbox{maximum interpretability-measure} \\ \mbox{source domain classifier} \end{array}$ & 362.83  & 5.44 & 451.93 & 0.9950 \\ \hline
              $\begin{array}{c} \mbox{maximum transferability-measure} \\ \mbox{transfer and multi-task learning} \end{array}$ & 362.83  & 5.44 & 451.93 & 0.9920    \\ \hdashline
       $\begin{array}{c} \mbox{maximum transferability-measure} \\ \mbox{source domain classifier} \end{array}$ & 362.83  & 5.44 & 451.93 & 0.9950 \\ \hline
        \hline 
    \end{tabular}  
  }  
\end{table}
Table~\ref{table_TAI_results_MNIST} reports the results obtained by the models that correspond to minimum privacy-leakage, maximum interpretability-measure, and maximum transferability-measure. The robustness of transfer and multi-task learning scenario is further highlighted in Table~\ref{table_TAI_results_MNIST}. To achieve the minimum value of privacy-leakage, the accuracy of source domain classifier must be decreased to 0.1760, however, the transfer and multi-task learning scenario achieves the minimum privacy-leakage value with the accuracy of 0.9510. As observed in Table~\ref{table_TAI_results_MNIST}, the maximum transferability-measure models also correspond to the maximum interpretability-measure models.  
\begin{figure}
\centering
\includegraphics[width = \textwidth]{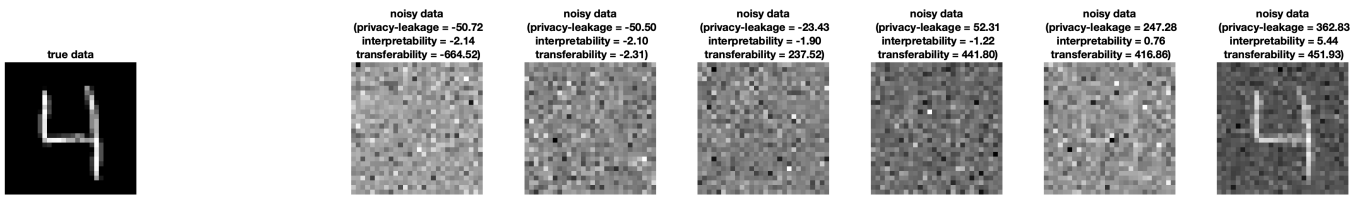}
\caption{An example of a source domain sample corresponding to different levels of privacy-leakage, interpretability-measure, and transferability-measure}
\label{fig-MNIST-TAI}
\end{figure}
As a visualization example, Fig.~\ref{fig-MNIST-TAI} displays noise added data samples for different values of information theoretic measures.

\subsection{Office and Caltech256 Datasets}
The ``Office+Caltech256'' dataset that has 10 common categories of both Office and Caltech256 datasets. The dataset has fours domains: \emph{amazon}, \emph{webcam}, \emph{dslr}, and \emph{caltech256}. This dataset has been widely used~\cite{Hoffman2013EfficientLO,Herath_2017_CVPR,8362683,Hoffman2014} for evaluating multi-class accuracy performance in a standard domain adaptation setting with a small number of labelled target samples. Following~\cite{Herath_2017_CVPR}, the 4096-dimensional deep-net VGG-FC6 features are extracted from the images. However for the learning of classifiers the 4096-dimensional feature vectors are reduced to 100-dimensional feature vectors using principal components computed from the data of \emph{amazon} domain. Thus, corresponding to each image, a 100-dimensional data vector is constructed.  
\paragraph{Interpretable Parameters:}
Corresponding to a data vector $y \in \mathbb{R}^{100}$, an interpretable parameter vector $t \in \{0,1\}^{10}$ is defined such that $j-$th element $t_j = 1$, if $j-$th class-label is associated to $y$, otherwise $t_j = 0$. That is, interpretable vector $t$, in our experimental setting, represents the class-label assigned to data vector $y$. 
\paragraph{Private Data:}
Here we assume that extracted image feature vectors are private, i.e., $x_{sr} = y_{sr}$.   
 \paragraph{Semi-Supervised Transfer Learning Scenario:}
Similarilly to~\cite{Hoffman2013EfficientLO,Herath_2017_CVPR,8362683,Hoffman2014}, the experimental setup is follows:    
\begin{enumerate}
\item the number of training samples per class in the source domain is 20 for \emph{amazon} and is 8 for other three domains,
\item the number of labelled samples per class in the target domain is 3 for all the four domains.
\end{enumerate}
\paragraph{Experimental Design:}
Taking a domain as source and another domain as target, 12 different transfer learning experiments are performed on the four domains associated to ``Office+Caltech256'' dataset. Each of the 12 experiments is repeated 20 times via creating 20 random train/test splits. In all of the 240 ($=12 \times 20$) experiments, Algorithm~\ref{algorithm_TAI} is applied three times with varying values of privacy-loss bound: first with differential privacy parameters as $(d=1,\epsilon = 0.01,\delta = 1\mathrm{e}{-5})$, second with differential privacy parameters as $(d=1,\epsilon = 0.1,\delta = 1\mathrm{e}{-5})$, and third with differential privacy parameters as $(d=1,\epsilon = 1,\delta = 1\mathrm{e}{-5})$. As Algorithm~\ref{algorithm_TAI} with different values of privacy-loss bound $\epsilon$ will result in different models, the transfer and multi-task learning models that correspond to maximum interpretability-measure and maximum transferability-measure are considered for an evaluation. 
\paragraph{Reference Methods:}
This dataset has been studied previously~\cite{Herath_2017_CVPR,7780918,Hoffman2014,6587717,Hoffman2013EfficientLO,8362683} and thus, as a reference, the performances of the following existing methods were considered:
\begin{enumerate}
\item ILS (1-NN)~\cite{Herath_2017_CVPR}: This method learns an Invariant Latent Space (ILS) to reduce the discrepancy between domains and uses Riemannian optimization techniques to match statistical properties between samples projected into the latent space from different domains. 
\item CDLS~\cite{7780918}: The Cross-Domain Landmark Selection (CDLS) method derives a domain-invariant feature subspace for heterogeneous domain adaptation. 
\item MMDT~\cite{Hoffman2014}: The Maximum Margin Domain Transform (MMDT) method adapts max-margin classifiers in a multi-class manner by learning a shared component of the domain shift as captured by the feature transformation. 
\item HFA~\cite{6587717}: The Heterogeneous Feature Augmentation (HFA) method learns common latent subspace and a classifier under max-margin framework.
\item OBTL~\cite{8362683}: The Optimal Bayesian Transfer Learning (OBTL) method employs Bayesian framework to transfer learning through modeling of a joint prior probability density function for feature-label distributions of the source and target domains.
\end{enumerate}
\paragraph{Results:}
Table~\ref{table_amazon_2_caltech}, Table~\ref{table_amazon_2_dslr}, Table~\ref{table_amazon_2_webcam}, Table~\ref{table_caltech_2_amazon}, Table~\ref{table_caltech_2_dslr}, Table~\ref{table_caltech_2_webcam}, Table~\ref{table_dslr_2_amazon}, Table~\ref{table_dslr_2_caltech}, Table~\ref{table_dslr_2_webcam}, Table~\ref{table_webcam_2_amazon}, Table~\ref{table_webcam_2_caltech}, and Table~\ref{table_webcam_2_dslr} report the results and the first two best performances have been marked. 
\begin{table}
\centering
 \caption{Accuracy (in \%, averaged over 20 experiments) obtained in \emph{amazon}$\rightarrow$\emph{caltech256} semi-supervised transfer learning experiments.}
 \label{table_amazon_2_caltech}
  {%
    \begin{tabular}{ccc}  
    \hline 
  \bfseries method & \bfseries feature type &  \bfseries accuracy (\%)  \\  
    \hline 
  $\begin{array}{c} \mbox{privacy-preserving maximum interpretability-measure model} \end{array}$   & VGG-FC6  & \underline{82.6}   \\
    $\begin{array}{c} \mbox{privacy-preserving maximum transferability-measure model} \end{array}$   & VGG-FC6  & \underline{82.6}   \\
non-private ILS (1-NN) & VGG-FC6 &  \underline{\textbf{83.3}}  \\
non-private CDLS & VGG-FC6 &  78.1  \\
non-private MMDT & VGG-FC6 & 78.7  \\
non-private HFA & VGG-FC6 &  75.5  \\
non-private OBTL & SURF &  41.5  \\
non-private ILS (1-NN) & SURF &  43.6  \\
non-private CDLS & SURF &  35.3  \\
non-private MMDT & SURF &  36.4 \\
non-private HFA & SURF & 31.0  \\
        \hline 
    \end{tabular}  
  }  
\end{table}   
\begin{table}
\centering
 \caption{Accuracy (in \%, averaged over 20 experiments) obtained in \emph{amazon}$\rightarrow$\emph{dslr} semi-supervised transfer learning experiments.}
 \label{table_amazon_2_dslr}
  {%
    \begin{tabular}{ccc}  
    \hline 
  \bfseries method & \bfseries feature type &  \bfseries accuracy (\%)  \\  
    \hline 
  $\begin{array}{c} \mbox{privacy-preserving maximum interpretability-measure model} \end{array}$   & VGG-FC6  & \underline{88.5}   \\
    $\begin{array}{c} \mbox{privacy-preserving maximum transferability-measure model} \end{array}$   & VGG-FC6  & \underline{\textbf{88.7}}   \\
non-private ILS (1-NN) & VGG-FC6 &  87.7 \\
non-private CDLS & VGG-FC6 &  86.9   \\
non-private MMDT & VGG-FC6 &  77.1   \\
non-private HFA & VGG-FC6 &  87.1   \\
non-private OBTL & SURF &  60.2   \\
non-private ILS (1-NN) & SURF &  49.8   \\
non-private CDLS & SURF &  60.4   \\
non-private MMDT & SURF &  56.7   \\
non-private HFA & SURF &  55.1  \\
        \hline 
    \end{tabular}  
  }  
\end{table}   
\begin{table}
\centering
 \caption{Accuracy (in \%, averaged over 20 experiments) obtained in \emph{amazon}$\rightarrow$\emph{webcam} semi-supervised transfer learning experiments.}
 \label{table_amazon_2_webcam}
  {%
    \begin{tabular}{ccc}  
    \hline 
  \bfseries method & \bfseries feature type & \bfseries accuracy (\%)  \\  
    \hline 
  $\begin{array}{c} \mbox{privacy-preserving maximum interpretability-measure model} \end{array}$   & VGG-FC6  & 89.3   \\
    $\begin{array}{c} \mbox{privacy-preserving maximum transferability-measure model} \end{array}$   & VGG-FC6  & 89.3   \\
non-private ILS (1-NN) & VGG-FC6 &  \underline{90.7}   \\
non-private CDLS & VGG-FC6 &  \underline{\textbf{91.2}}  \\
non-private MMDT & VGG-FC6 &  82.5   \\
non-private HFA & VGG-FC6 &  87.9   \\
non-private OBTL & SURF &  72.4   \\
non-private ILS (1-NN) & SURF &  59.7   \\
non-private CDLS & SURF &  68.7   \\
non-private MMDT & SURF &   64.6  \\
non-private HFA & SURF &  57.4  \\
        \hline 
    \end{tabular}  
  }  
\end{table}   
\begin{table}
\centering
 \caption{Accuracy (in \%, averaged over 20 experiments) obtained in \emph{caltech256}$\rightarrow$\emph{amazon} semi-supervised transfer learning experiments.}
 \label{table_caltech_2_amazon}
  {%
    \begin{tabular}{ccc}  
    \hline 
  \bfseries method & \bfseries feature type &  \bfseries accuracy (\%)  \\  
    \hline 
  $\begin{array}{c} \mbox{privacy-preserving maximum interpretability-measure model} \end{array}$   & VGG-FC6  & \underline{\textbf{92.6}}  \\
    $\begin{array}{c} \mbox{privacy-preserving maximum transferability-measure model} \end{array}$   & VGG-FC6  & \underline{\textbf{92.6}}   \\
non-private ILS (1-NN) & VGG-FC6 &  \underline{89.7}   \\
non-private CDLS & VGG-FC6 &  88.0  \\
non-private MMDT & VGG-FC6 &  85.9  \\
non-private HFA & VGG-FC6 &  86.2   \\
non-private OBTL & SURF &   54.8  \\
non-private ILS (1-NN) & SURF &   55.1   \\
non-private CDLS & SURF &   50.9  \\
non-private MMDT & SURF &  49.4   \\
non-private HFA & SURF &  43.8  \\
        \hline 
    \end{tabular}  
  }  
\end{table}   
\begin{table}
\centering
 \caption{Accuracy (in \%, averaged over 20 experiments) obtained in \emph{caltech256}$\rightarrow$\emph{dslr} semi-supervised transfer learning experiments.}
 \label{table_caltech_2_dslr}
  {%
    \begin{tabular}{ccc}  
    \hline 
  \bfseries method & \bfseries feature type &  \bfseries accuracy (\%)  \\  
    \hline 
  $\begin{array}{c} \mbox{privacy-preserving maximum interpretability-measure model} \end{array}$   & VGG-FC6  & \underline{\textbf{89.1}}   \\
    $\begin{array}{c} \mbox{privacy-preserving maximum transferability-measure model} \end{array}$   & VGG-FC6  & \underline{\textbf{89.1}}   \\
non-private ILS (1-NN) & VGG-FC6 &   86.9  \\
non-private CDLS & VGG-FC6 &  86.3  \\
non-private MMDT & VGG-FC6 &   77.9   \\
non-private HFA & VGG-FC6 &  87.0    \\
non-private OBTL & SURF &  61.5   \\
non-private ILS (1-NN) & SURF &   56.2  \\
non-private CDLS & SURF &   59.8  \\
non-private MMDT & SURF &   56.5  \\
non-private HFA & SURF &   55.6 \\
        \hline 
    \end{tabular}  
  }  
\end{table}   
\begin{table}
\centering
 \caption{Accuracy (in \%, averaged over 20 experiments) obtained in \emph{caltech256}$\rightarrow$\emph{webcam} semi-supervised transfer learning experiments.}
 \label{table_caltech_2_webcam}
  {%
    \begin{tabular}{ccc}  
    \hline 
  \bfseries method & \bfseries feature type &  \bfseries accuracy (\%)  \\  
    \hline 
  $\begin{array}{c} \mbox{privacy-preserving maximum interpretability-measure model} \end{array}$   & VGG-FC6  & 87.8   \\
    $\begin{array}{c} \mbox{privacy-preserving maximum transferability-measure model} \end{array}$   & VGG-FC6  &87.7   \\
non-private ILS (1-NN) & VGG-FC6 &  \underline{\textbf{91.4}}   \\
non-private CDLS & VGG-FC6 &  \underline{89.7}   \\
non-private MMDT & VGG-FC6 &  82.8   \\
non-private HFA & VGG-FC6 &  86.0    \\
non-private OBTL & SURF &  71.1   \\
non-private ILS (1-NN) & SURF &   62.9  \\
non-private CDLS & SURF &  66.3   \\
non-private MMDT & SURF &   63.8  \\
non-private HFA & SURF &  58.1  \\
        \hline 
    \end{tabular}  
  }  
\end{table}   
\begin{table}
\centering
 \caption{Accuracy (in \%, averaged over 20 experiments) obtained in \emph{dslr}$\rightarrow$\emph{amazon} semi-supervised transfer learning experiments.}
 \label{table_dslr_2_amazon}
  {%
    \begin{tabular}{ccc}  
    \hline 
  \bfseries method & \bfseries feature type &  \bfseries accuracy (\%)   \\  
    \hline 
  $\begin{array}{c} \mbox{privacy-preserving maximum interpretability-measure model} \end{array}$   & VGG-FC6  & \underline{\textbf{91.9}}    \\
    $\begin{array}{c} \mbox{privacy-preserving maximum transferability-measure model} \end{array}$   & VGG-FC6  & \underline{\textbf{91.9}}    \\
non-private ILS (1-NN) & VGG-FC6 &  \underline{88.7}   \\
non-private CDLS & VGG-FC6 &   88.1  \\
non-private MMDT & VGG-FC6 &   83.6  \\
non-private HFA & VGG-FC6 &   85.9   \\
non-private OBTL & SURF &  54.4  \\
non-private ILS (1-NN) & SURF &   55.0  \\
non-private CDLS & SURF &  50.7  \\
non-private MMDT & SURF &  46.9  \\
non-private HFA & SURF &   42.9  \\
        \hline 
    \end{tabular}  
  }  
\end{table}   
 \begin{table}
\centering
 \caption{Accuracy (in \%, averaged over 20 experiments) obtained in \emph{dslr}$\rightarrow$\emph{caltech256} semi-supervised transfer learning experiments.}
 \label{table_dslr_2_caltech}
  {%
    \begin{tabular}{ccc}  
    \hline 
  \bfseries method & \bfseries feature type & \bfseries accuracy (\%)  \\  
    \hline 
  $\begin{array}{c} \mbox{privacy-preserving maximum interpretability-measure model} \end{array}$   & VGG-FC6  & \underline{\textbf{82.9}}   \\
    $\begin{array}{c} \mbox{privacy-preserving maximum transferability-measure model} \end{array}$   & VGG-FC6  & \underline{\textbf{82.9}}   \\
non-private ILS (1-NN) & VGG-FC6 &   \underline{81.4}  \\
non-private CDLS & VGG-FC6 &  77.9   \\
non-private MMDT & VGG-FC6 &  71.8  \\
non-private HFA & VGG-FC6 &  74.8   \\
non-private OBTL & SURF &  40.3  \\
non-private ILS (1-NN) & SURF &  41.0  \\
non-private CDLS & SURF &  34.9  \\
non-private MMDT & SURF &  34.1  \\
non-private HFA & SURF &   30.9  \\
        \hline 
    \end{tabular}  
  }  
\end{table}   
 \begin{table}
\centering
 \caption{Accuracy (in \%, averaged over 20 experiments) obtained in \emph{dslr}$\rightarrow$\emph{webcam} semi-supervised transfer learning experiments.}
 \label{table_dslr_2_webcam}
  {%
    \begin{tabular}{cccc}  
    \hline 
  \bfseries method & \bfseries feature type &  \bfseries accuracy (\%)  \\  
    \hline 
  $\begin{array}{c} \mbox{privacy-preserving maximum interpretability-measure model} \end{array}$   & VGG-FC6  & 88.9  \\
    $\begin{array}{c} \mbox{privacy-preserving maximum transferability-measure model} \end{array}$   & VGG-FC6  & 89.0   \\
non-private ILS (1-NN) & VGG-FC6 &   \underline{\textbf{95.5}}  \\
non-private CDLS & VGG-FC6 &  \underline{90.7}  \\
non-private MMDT & VGG-FC6 &   86.1 \\
non-private HFA & VGG-FC6 &   86.9  \\
non-private OBTL & SURF &  83.2 \\
non-private ILS (1-NN) & SURF &  80.1   \\
non-private CDLS & SURF &  68.5  \\
non-private MMDT & SURF &  74.1  \\
non-private HFA & SURF &  60.5   \\
        \hline 
    \end{tabular}  
  }  
\end{table}   
\begin{table}
\centering
 \caption{Accuracy (in \%, averaged over 20 experiments) obtained in \emph{webcam}$\rightarrow$\emph{amazon} semi-supervised transfer learning experiments.}
 \label{table_webcam_2_amazon}
  {%
    \begin{tabular}{cccc}  
    \hline 
  \bfseries method & \bfseries feature type &  \bfseries accuracy (\%)  \\  
    \hline 
  $\begin{array}{c} \mbox{privacy-preserving maximum interpretability-measure model} \end{array}$   & VGG-FC6  & \underline{\textbf{92.3}}   \\
    $\begin{array}{c} \mbox{privacy-preserving maximum transferability-measure model} \end{array}$   & VGG-FC6  & \underline{\textbf{92.3}}   \\
non-private ILS (1-NN) & VGG-FC6 &  \underline{88.8}   \\
non-private CDLS & VGG-FC6 &  87.4  \\
non-private MMDT & VGG-FC6 &  84.7  \\
non-private HFA & VGG-FC6 &  85.1  \\
non-private OBTL & SURF &   55.0 \\
non-private ILS (1-NN) & SURF & 54.3   \\
non-private CDLS & SURF &  51.8  \\
non-private MMDT & SURF &  47.7  \\
non-private HFA & SURF &  56.5   \\
        \hline 
    \end{tabular}  
  }  
\end{table}   
 \begin{table}
\centering
 \caption{Accuracy (in \%, averaged over 20 experiments) obtained in \emph{webcam}$\rightarrow$\emph{caltech256} semi-supervised transfer learning experiments.}
 \label{table_webcam_2_caltech}
  {%
    \begin{tabular}{cccc}  
    \hline 
  \bfseries method & \bfseries feature type &  \bfseries accuracy (\%)  \\  
    \hline 
  $\begin{array}{c} \mbox{privacy-preserving maximum interpretability-measure model} \end{array}$   & VGG-FC6  & \underline{81.4}   \\
    $\begin{array}{c} \mbox{privacy-preserving maximum transferability-measure model} \end{array}$   & VGG-FC6  & \underline{81.4}   \\
non-private ILS (1-NN) & VGG-FC6 &  \underline{\textbf{82.8}}   \\
non-private CDLS & VGG-FC6 &  78.2  \\
non-private MMDT & VGG-FC6 &  73.6  \\
non-private HFA & VGG-FC6 &  74.4   \\
non-private OBTL & SURF &  37.4   \\
non-private ILS (1-NN) & SURF &   38.6   \\
non-private CDLS & SURF &  33.5   \\
non-private MMDT & SURF & 32.2   \\
non-private HFA & SURF &   29.0  \\
        \hline 
    \end{tabular}  
  }  
\end{table}
\begin{table}
\centering
 \caption{Accuracy (in \%, averaged over 20 experiments) obtained in \emph{webcam}$\rightarrow$\emph{dslr} semi-supervised transfer learning experiments.}
 \label{table_webcam_2_dslr}
  {%
    \begin{tabular}{ccc}  
    \hline 
  \bfseries method & \bfseries feature type &  \bfseries accuracy (\%)  \\  
    \hline 
  $\begin{array}{c} \mbox{privacy-preserving maximum interpretability-measure model} \end{array}$   & VGG-FC6  & \underline{90.8}   \\
    $\begin{array}{c} \mbox{privacy-preserving maximum transferability-measure model} \end{array}$   & VGG-FC6  & 90.2   \\
non-private ILS (1-NN) & VGG-FC6 &  \underline{\textbf{94.5}}  \\
non-private CDLS & VGG-FC6 &  88.5   \\
non-private MMDT & VGG-FC6 &   85.1  \\
non-private HFA & VGG-FC6 &   87.3 \\
non-private OBTL & SURF &  75.0   \\
non-private ILS (1-NN) & SURF  & 70.8    \\
non-private CDLS & SURF &  60.7  \\
non-private MMDT & SURF &   67.0  \\
non-private HFA & SURF &   56.5  \\
        \hline 
    \end{tabular}  
  }  
\end{table}
\begin{table}
\centering
\caption{Comparison of the methods on ``Office+Caltech256'' dataset.}
 \label{table_overall_performance}
 {
   \begin{tabular}{ccc}  
     \hline 
  \bfseries method &  \bfseries $\begin{array}{c} \mbox{number of experiments} \\ \mbox{in which method} \\ \mbox{performed best} \end{array}$     \\  
    \hline 
    $\begin{array}{c} \mbox{privacy-preserving maximum transferability-measure model} \end{array}$  & 6  \\ 
        $\begin{array}{c} \mbox{privacy-preserving maximum interpretability-measure model} \end{array}$  & 5  \\ 
  non-private ILS (1-NN) &  5  \\ 
  non-private CDLS & 1  \\
 \hline
   \end{tabular}
 }
\end{table}
Finally, Table~\ref{table_overall_performance} summarizes the overall performance of top four methods. As observed in Table~\ref{table_overall_performance}, the maximum transferability-measure model remains as best performing in maximum number of experiments. The most remarkable result observed is that the proposed methodology, despite being privacy-preserving ensuring differential privacy-loss bound to be less than equal to 1 and not requiring an access to source data samples, performs better than even the non-private methods.   
\subsection{An Application Example: Mental Stress Detection}
\begin{figure}
\centerline{ \subfigure[minimum privacy-leakage models]{\includegraphics[width=0.5\textwidth]{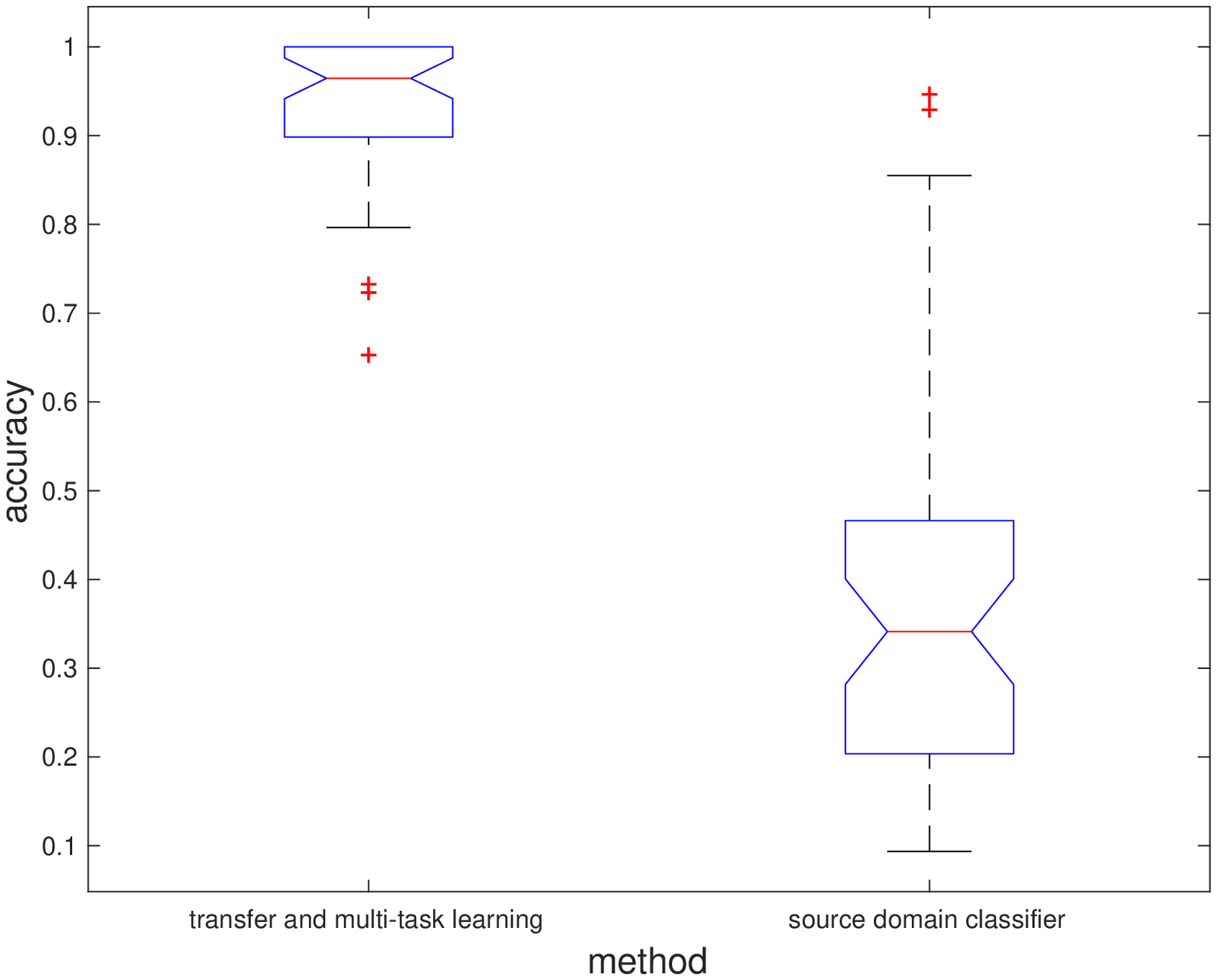}\label{fig-min-privacy-leakage-stress}} \hfil  \subfigure[maximum interpretability-measure models]{\includegraphics[width=0.5\textwidth]{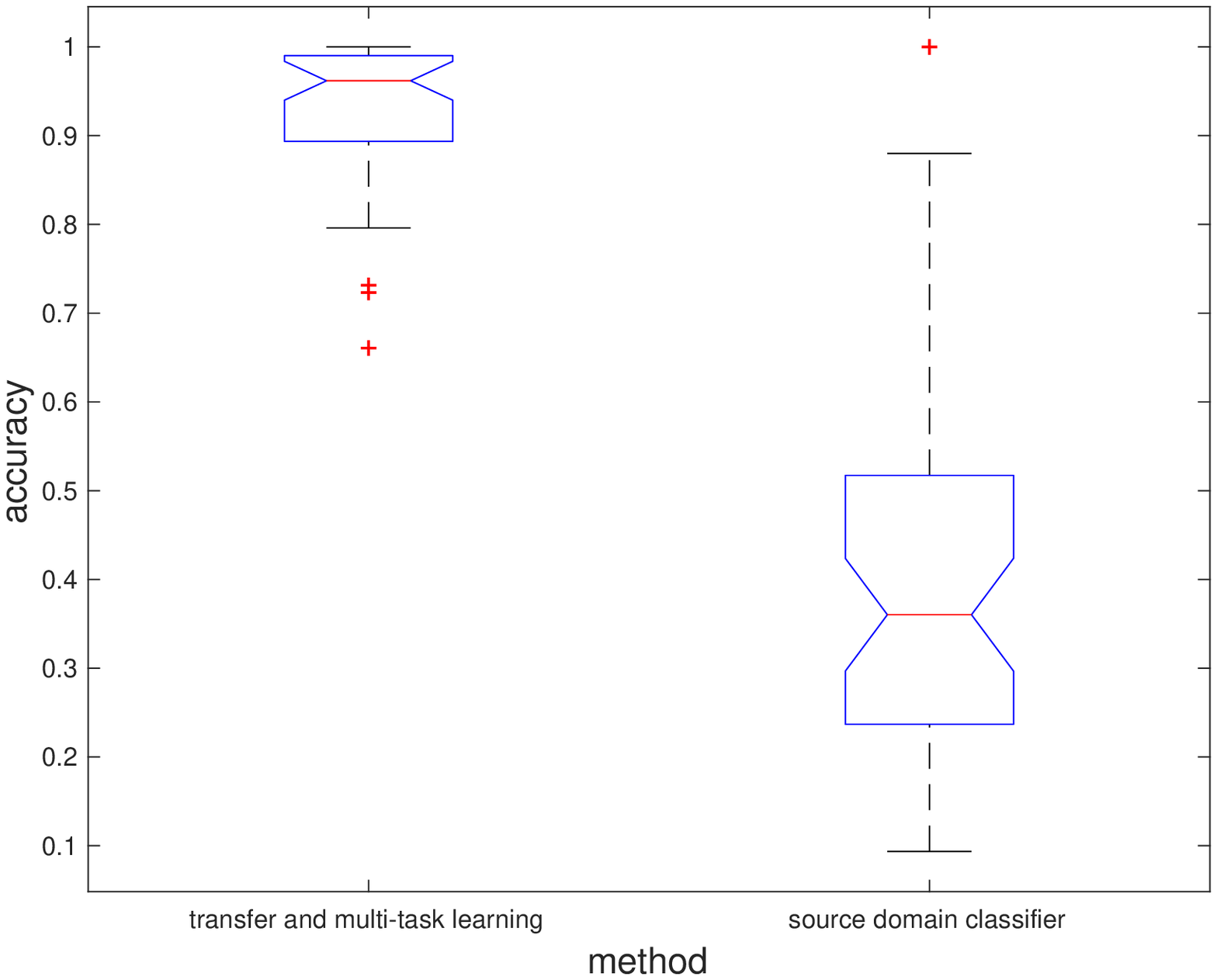}\label{fig-max-interpretability-stress}}}
\caption{The box-plots of accuracies obtained in detecting mental stress on 48 different subjects.}
\label{fig_TAI_results_stress}
\end{figure}

The mental stress detection problem is considered as an application example of the proposed privacy-preserving interpretable and transferable learning approach. The dataset from~\cite{9216097}, consisting of heart rate interval measurements of different subjects, is considered for the study of individual stress detection problem. In~\cite{9216097}, a membership-mappings based interpretable deep model was applied for an estimation of stress-score, however, current study deals with application of the proposed privacy-preserving interpretable and transferable deep learning method to solve stress classification problem. The problem is concerned with the detection of stress on an individual based on the analysis of recorded sequence of R-R intervals, $\{RR^i\}_i$. The R-R data vector at $i-$th time-index, $y^i$, is defined as
\begin{IEEEeqnarray}{rCl}
 y^i & = & \left[\begin{IEEEeqnarraybox*}[][c]{,c/c/c/c,} RR^i & RR^{i-1} & \cdots & RR^{i-d}\end{IEEEeqnarraybox*} \right]^T.
  \end{IEEEeqnarray}     
That is, the current interval and history of previous $d$ intervals constitute the data vector. Assuming an average heartbeat of 72 beats per minute, $d$ is chosen as equal to $72 \times 3 = 216$ so that R-R data vector consists of on an average 3-minutes long R-R intervals sequence. A dataset, say $\{y^i\}_{i}$, is built via 1) preprocessing the R-R interval sequence $\{RR^i\}_i$ with an impulse rejection filter~\cite{1403112} for artifacts detection, and 2) excluding the R-R data vectors containing artifacts from the dataset. The dataset contains the stress-score on a scale from 0 to 100. A label of either ``\emph{no-stress}'' or ``\emph{under-stress}'' is assigned to each $y^i$ based on the stress-score. Thus, we have a binary classification problem.     
\paragraph{Interpretable Parameters:}
Corresponding to a R-R data vector, there exists the set of interpretable parameters: \emph{mental demand}, \emph{physical demand}, \emph{temporal demand}, \emph{own performance}, \emph{effort}, and \emph{frustration}. These are the six components of stress acquired using NASA Task Load Index~\cite{hart1988development}. NASA Task Load Index provides subjective assessment of stress where an individual provides a rating on the scale from 0 to 100 for each of the six components of stress (mental demand, physical demand, temporal demand, own performance, effort, and frustration). Thus corresponding to each $217-$dimensional R-R data vector, there exists a 6-dimensional interpretable parameters vector acquired using NASA Task Load Index.
\paragraph{Private Data:}
Here we assume that heart rate values are private. As instantaneous heart rate is given as $HR^i = 60/RR^i$, thus an information about private data is directly contained in the R-R data vectors.   
\paragraph{Semi-Supervised Transfer Learning Scenario:}
Out of total subjects, a randomly chosen subject's data serve as the source domain data. Considering every other subject's data as the target domain data, the transfer learning experiment is performed independently on each target subject where 50\% of the target subject's samples are labelled and remaining unlabelled target samples also serve as test data for evaluating the classification performance. However, only the target subjects, with data containing both the classes and at least 60 samples, were considered for experimentation. There are in total 48 such target subjects.          
\paragraph{Experimental Design:}
Algorithm~\ref{algorithm_TAI} is applied with the differential privacy parameters as $d=1$, $\epsilon \in \{0.1,0.5,1,2,5,8,20,50,100,\infty\}$, and $\delta = 1\mathrm{e}{-5}$. Each of 48 experiments involves 10 different privacy-preserving semi-supervised transfer learning scenarios with privacy-loss bound values as $\epsilon = 0.1$, $\epsilon = 0.5$, $\epsilon = 1$, $\epsilon = 2$, $\epsilon = 5$, $\epsilon = 8$, $\epsilon = 20$, $\epsilon = 50$, $\epsilon=100$, and $\epsilon = \infty$. There are following two requirements associated to this application example:
\begin{enumerate}
\item the private source domain data must be protected while transferring knowledge from source to target domain, and
\item the interpretability of the source domain model should be high.    
\end{enumerate}   
In view of the aforementioned requirements, the models, that correspond to minimum privacy-leakage and maximum interpretability-measure amongst all the models obtained corresponding to 10 different choices of differential privacy-loss bound $\epsilon$, are considered for detecting stress.  
\paragraph{Results:}
Fig.~\ref{fig_TAI_results_stress} summarizes the experimental results where accuracies obtained by both minimum privacy-leakage models and maximum interpretability-measure models have been displayed as box-plots.   
\begin{table}
\centering
 \caption{Results (median values) obtained in stress detection experiments on a dataset consisting of heart rate interval measurements.}
 \label{table_TAI_results_stress}
  {%
    \begin{tabular}{ccccc}  
    \hline 
    \bfseries Method & \bfseries  $\begin{array}{c} \mbox{privacy-} \\ \mbox{leakage} \end{array}$  & \bfseries $\begin{array}{c} \mbox{interpretability-} \\ \mbox{measure} \end{array}$ & \bfseries $\begin{array}{c} \mbox{transferability-} \\ \mbox{measure} \end{array}$ & \bfseries $\begin{array}{c} \mbox{classification} \\ \mbox{accuracy} \end{array}$   \\  
    \hline 
       $\begin{array}{c} \mbox{minimum privacy-leakage} \\ \mbox{transfer and multi-task learning} \end{array}$ & -3.74   & 3.47  & 291.84  & 0.9647 \\ \hdashline
       $\begin{array}{c} \mbox{minimum privacy-leakage} \\ \mbox{source domain classifier} \end{array}$ & -3.74  & 3.47  & 291.84 & 0.3411 \\ \hline
       $\begin{array}{c} \mbox{maximum interpretability-measure} \\ \mbox{transfer and multi-task learning} \end{array}$ & 0.43   & 23.92  & 773.36  & 0.9619    \\ \hdashline
       $\begin{array}{c} \mbox{maximum interpretability-measure} \\ \mbox{source domain classifier} \end{array}$ & 0.43   & 23.92  & 773.36  & 0.3602  \\ 
        \hline
        \hline 
    \end{tabular}  
  }  
\end{table}
It is observed in Fig.~\ref{fig_TAI_results_stress} that the transfer and multi-task learning improves considerably the performance of source domain classifier. Table~\ref{table_TAI_results_stress} reports the median values (of privacy-leakage, interpretability-measure, transferability-measure, and classification accuracy) obtained in the experiments on 48 different subjects. The robust performance of transfer and multi-task learning scenario is further observed in Table~\ref{table_TAI_results_stress}.    
\begin{figure}
\centering
\includegraphics[width = \textwidth]{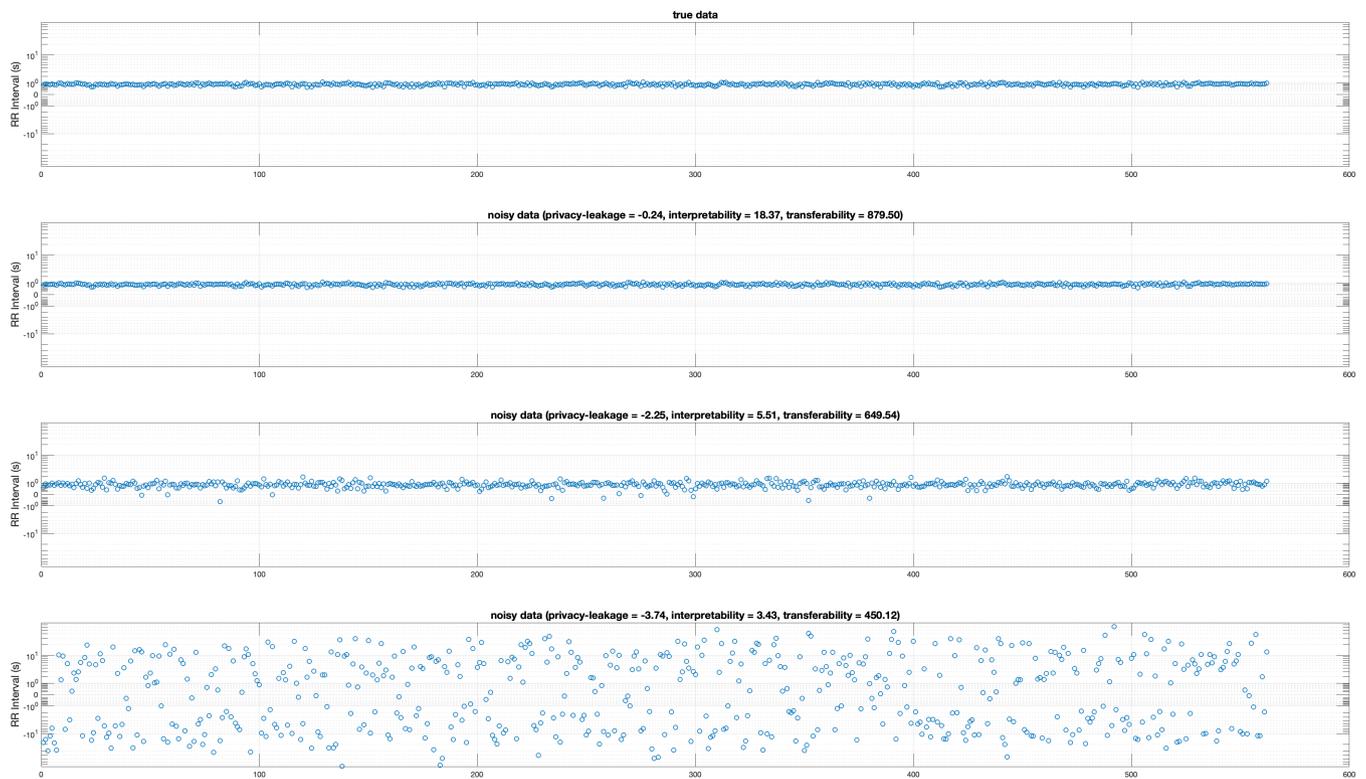}
\caption{A display of source domain R-R interval data corresponding to different levels of privacy-leakage, interpretability-measure, and transferability-measure}
\label{fig-demo-HR-data}
\end{figure}
As a visualization example, Fig.~\ref{fig-demo-HR-data} displays the noise added source domain heart rate interval data for different values of information theoretic measures.
\clearpage
\section{Concluding Remarks}\label{sec_conclusion}      
The paper has introduced an information theoretic trustworthy AI framework. The information theoretic measures have been defined for privacy-leakage, interpretability, and transferability to study the tradeoffs. This is the first study to develop information theory based unified approach to trustworthy AI. Although the text has not focused on federated and distributed learning, the transfer learning approach could be easily extended to the multi-party system and the transferability-measure could be calculated for any pair of parties. Also, the explainability of the conditionally deep autoencoders follows, similar to as in~\cite{9216097}, via estimating interpretable parameters from non-interpretable data feature vectors using variational membership-mapping Bayesian model. Further, the variational membership-mapping Bayesian model quantifies uncertainties on the estimation of parameters (of interest) which is also important for a user's trust on the model. The considered unified approach to privacy-preserving interpretable and transferable learning involves membership-mappings based conditionally deep autoencoders, albeit other data representation learning models could be explored under the proposed trustworthy AI framework.

\section*{Acknowledgments}
The research reported in this paper has been partly supported by Supported by the Austrian Research Promotion Agency (FFG) Sub-Project PETAI (Privacy Secured Explainable and Transferable AI for Healthcare Systems); the Federal Ministry for Climate Action, Environment, Energy, Mobility, Innovation and Technology (BMK); the Federal Ministry for Digital and Economic Affairs (BMDW); and the Province of Upper Austria in the frame of the COMET - Competence Centers for Excellent Technologies Programme managed by Austrian Research Promotion Agency FFG.

\appendix
\section{Algorithms}\label{appendix_algorithms}
\begin{algorithm}
\caption{Variational learning of the membership-mappings~\cite{kumar2022differentially}}
\label{algorithm_basic_learning}
\begin{algorithmic}[1]
\Require  Dataset $\left\{ (x^i,y^i) \; | \; x^i \in \mathbb{R}^n,\; y^i \in \mathbb{R}^p,\; i \in \{1,\cdots,N \} \right \}$ and maximum possible number of auxiliary points $M_{max} \in \mathbb{Z}_+$ with $M_{max} \leq N$.  
\State Choose $\nu$ and $w = (w_1,\cdots,w_n)$ as in (\ref{eq_738496.4701}) and (\ref{eq_738496.4698}) respectively.  
\State Choose a small positive value $\kappa = 10^{-1}$. 
\State Set iteration count $it = 0$ and $M|_0 = M_{max}$.
\While{$\tau(M|_{it},1) < \kappa$}
\State $M|_{it+1} = \lceil 0.9M|_{it} \rceil$
\State $it \leftarrow it + 1$
\EndWhile
\State Set $M = M|_{it}$.
\If{$\tau(M,1) \geq  \frac{1}{p} \sum_{j=1}^p \text{var}\left(y_j^1,\cdots,y_j^N\right)$}
\State $\sigma^2 = 1$
\Else 
\State $\sigma^2 = \frac{1}{\tau(M,1)} \frac{1}{p} \sum_{j=1}^p \text{var}\left(y_j^1,\cdots,y_j^N\right)$
\EndIf
\State Compute $\mathrm{a} = \{ a^{m}\}_{m=1}^M$ using (\ref{eq_738496.4692}), $K_{\mathrm{x}\mathrm{x}}$ using (\ref{738026.844153}), $K_{\mathrm{a}\mathrm{a}}$ using (\ref{eq_membership1004_2}), and $K_{\mathrm{x}\mathrm{a}}$ using (\ref{eq_738497.4922}).
\State Set $\beta = 1$.
 \Repeat
\State Compute $\alpha$ using (\ref{eq_vector_alpha}).
\State Update the value of $\beta$ using (\ref{eq_738497.4473}).
 \Until{($\beta$ nearly converges)}
\State Compute $\alpha$ using (\ref{eq_vector_alpha}).
\State \Return the parameters set $\mathbb{M} = \{\alpha, \mathrm{a}, M,\sigma,w\}$.
\end{algorithmic} 
\end{algorithm} 
With reference to Algorithm~\ref{algorithm_basic_learning}, we have followings:
\begin{itemize}
\item The degrees of freedom associated to the Student-t membership-mapping $\nu \in \mathbb{R}_{+} \setminus [0,2]$ is chosen as 
\begin{IEEEeqnarray}{rCl}
\label{eq_738496.4701}\nu & = & 2.1
 \end{IEEEeqnarray} 
 \item The auxiliary inducing points are suggested to be chosen as the cluster centroids: 
\begin{IEEEeqnarray}{rCl}
\label{eq_738496.4692}\mathrm{a} = \{ a^{m}\}_{m=1}^M  =  cluster\_centroid(  \{x^i\}_{i=1}^N, M) 
 \end{IEEEeqnarray} 
where $cluster\_centroid(  \{ x^i \}_{i=1}^N,M)$ represents the k-means clustering on $ \{ x^i \}_{i=1}^N$. 
 \item The parameters $(w_1,\cdots,w_n)$ for kernel function~(\ref{eq_membership1003_3}) are chosen such that $w_{k}$ (for $k\in \{1,2,\cdots,n\}$) is given as
\begin{IEEEeqnarray}{rCl}
\label{eq_738496.4698}w_k & = & \left(\max_{1 \leq i \leq N}\left(x^i_k\right) - \min_{1 \leq i \leq N}\left(x^i_k\right)\right)^{-2}
 \end{IEEEeqnarray} 
where $x^i_k$ is the $k-$th element of vector $x^i \in \mathbb{R}^n$.
\item $K_{\mathrm{a}\mathrm{a}} \in \mathbb{R}^{M \times M}$ and $K_{\mathrm{x}\mathrm{a}}  \in \mathbb{R}^{N \times M}$ are matrices with their $(i,j)-$th elements given as
\begin{IEEEeqnarray}{rCl}
\label{eq_membership1004_2} \left( K_{\mathrm{a}\mathrm{a}} \right)_{i,j} & = & kr(a^{i},a^{j}) \\
\label{eq_738497.4922} \left( K_{\mathrm{x}\mathrm{a}} \right)_{i,j} & = & kr(x^{i},a^{j})
 \end{IEEEeqnarray} 
where $kr: \mathbb{R}^n \times \mathbb{R}^n \rightarrow \mathbb{R}$ is a positive definite kernel function defined as in (\ref{eq_membership1003_3}).
\item The scalar-valued function $\tau(M,\sigma^2)$ is defined as
\begin{IEEEeqnarray}{rCl}
\tau(M,\sigma^2) & := & \frac{Tr(K_{\mathrm{x}\mathrm{x}}) - Tr((K_{\mathrm{a}\mathrm{a}})^{-1}  K_{\mathrm{x}\mathrm{a}}^T K_{\mathrm{x}\mathrm{a}} )}{\nu+ M - 2}
 \end{IEEEeqnarray} 
where $\mathrm{a}$ is given by (\ref{eq_738496.4692}), $\nu$ is given by (\ref{eq_738496.4701}), and parameters $(w_1,\cdots,w_n)$ (which are required to evaluate the kernel function for computing matrices $K_{\mathrm{x}\mathrm{x}}$, $K_{\mathrm{a}\mathrm{a}}$, and $K_{\mathrm{x}\mathrm{a}}$) are given by (\ref{eq_738496.4698}).
\item $\alpha = \left[\begin{IEEEeqnarraybox*}[][c]{,c/c/c,}  \alpha_1 & \cdots & \alpha_p
 \end{IEEEeqnarraybox*} \right] \in \mathbb{R}^{M \times p}$ is a matrix with its $j-$th column defined as 
\begin{IEEEeqnarray}{rCl}
\label{eq_vector_alpha}  \alpha_j & := & \left( K_{\mathrm{x}\mathrm{a}}^T K_{\mathrm{x}\mathrm{a}}   + \frac{Tr(K_{\mathrm{x}\mathrm{x}}) - Tr((K_{\mathrm{a}\mathrm{a}})^{-1}  K_{\mathrm{x}\mathrm{a}}^T K_{\mathrm{x}\mathrm{a}} )}{\nu+ M - 2}  K_{\mathrm{a}\mathrm{a}}   +    \frac{K_{\mathrm{a}\mathrm{a}}}{\beta }\right)^{-1}  (K_{\mathrm{x}\mathrm{a}} )^T \mathrm{y}_j  
  \end{IEEEeqnarray}
  \item The disturbance precision value $\beta$ is iteratively estimated as 
\begin{IEEEeqnarray}{rCl}
\label{eq_738497.4473} \frac{1}{\beta} & = & \frac{1}{pN}\sum_{j=1}^p \sum_{i=1}^N \left |y_j^i -  \widehat{ \mathcal{F}_j(x^{i})} \right |^2
 \end{IEEEeqnarray} 
where $\widehat{ \mathcal{F}_j(x^{i})}$ is the estimated membership-mapping output given as 
\begin{IEEEeqnarray}{rCl}
\label{eq_final_layer_output} \widehat{ \mathcal{F}_j(x^{i})} & = & \left(G(x^i) \right) \alpha_j.
  \end{IEEEeqnarray}   
Here, $G(x) \in \mathbb{R}^{1 \times M}$ is a vector-valued function defined as
\begin{IEEEeqnarray}{rCl}
\label{eq_738495.5497} G(x)& := &  \left[\begin{IEEEeqnarraybox*}[][c]{,c/c/c,}kr(x,a^{1}) & \cdots & kr(x,a^{M}) \end{IEEEeqnarraybox*} \right]
 \end{IEEEeqnarray} 
where $kr: \mathbb{R}^n \times \mathbb{R}^n \rightarrow \mathbb{R}$ is defined as in (\ref{eq_membership1003_3}).
\end{itemize} 
\begin{algorithm}
\caption{Variational learning of CDMMA~\cite{10.1007/978-3-030-87101-7_14,kumar2022differentially}}
\label{algorithm_DSFMA}
\begin{algorithmic}[1]
\Require Data set $\mathbf{Y} = \left\{ y^i \in \mathbb{R}^p \; | \; i \in \{1,\cdots,N \} \right \}$; the subspace dimension $n \in \{1,2,\cdots,p \}$; maximum number of auxiliary points $M_{max} \in \mathbb{Z}_+$ with $M_{max} \leq N$; the number of layers $L \in \mathbb{Z}_{+}$.
\For{$l=1$ to $L$}
\State Set subspace dimension associated to $l-$th layer as $n_l = \max(n - l + 1,1)$.
\State Define $P^l \in \mathbb{R}^{n_l \times p}$ such that $i-$th row of $P^l$ is equal to transpose of eigenvector corresponding to $i-$th largest eigenvalue of sample covariance matrix of data set $\mathbf{Y} $. 
\State Define a latent variable $x^{l,i} \in \mathbb{R}^{n_l}$, for $i \in \{1,\cdots,N \}$, as
 \begin{IEEEeqnarray}{rCl}
\label{eq_x_l_i}x^{l,i} &:=& \left\{ \,
    \begin{IEEEeqnarraybox}[][c]{l?s}
      \IEEEstrut
      P^ly^i & if $l=1$, \\
     P^l \hat{y}^{l-1}(x^{l-1,i};\mathbb{M}^{l-1}) & if $l > 1$
      \IEEEstrut
    \end{IEEEeqnarraybox}
\right.  \IEEEeqnarraynumspace
\end{IEEEeqnarray}   
where $\hat{y}^{l-1}$ is the estimated output of the $(l-1)-$th layer computed using (\ref{eq_738124.770095}) for the parameters set $\mathbb{M}^{l-1} = \{\alpha^{l-1}, \mathrm{a}^{l-1}, M^{l-1}, \sigma^{l-1},w^{l-1}\}$.  
\State Define $M_{max}^l$ as
 \begin{IEEEeqnarray}{rCl}
\label{eq_738499.4927}M_{max}^l &:=& \left\{ \,
    \begin{IEEEeqnarraybox}[][c]{l?s}
      \IEEEstrut
      M_{max} & if $l=1$, \\
     M^{l-1} & if $l > 1$
      \IEEEstrut
    \end{IEEEeqnarraybox}
\right.  \IEEEeqnarraynumspace
\end{IEEEeqnarray} 
\State Compute parameters set $\mathbb{M}^l = \{\alpha^{l}, \mathrm{a}^{l}, M^{l}, \sigma^{l},w^{l}\}$, characterizing the membership-mappings associated to $l-$th layer, using Algorithm~\ref{algorithm_basic_learning} on data set $\left\{ (x^{l,i},y^i) \; | \;  i \in \{1,\cdots,N \} \right \}$ with maximum possible number of auxiliary points $M_{max}^l$. 
\EndFor
\State \Return the parameters set $\mathcal{M} = \{\{\mathbb{M}^1,\cdots,\mathbb{M}^L\}, \{P^1,\cdots,P^L \} \}$.
\end{algorithmic}
\end{algorithm} 
\begin{algorithm}
\caption{Variational learning of wide CDMMA~\cite{10.1007/978-3-030-87101-7_14,kumar2022differentially}}
\label{algorithm_WDSFMA}
\begin{algorithmic}[1]
\Require  Data set $\mathbf{Y} = \left\{ y^i \in \mathbb{R}^p \; | \; i \in \{1,\cdots,N \} \right \}$; the subspace dimension $n \in \{1,2,\cdots,p\}$; ratio $r_{max} \in (0,1]$; the number of layers $L \in \mathbb{Z}_{+}$.
\State Apply k-means clustering to partition $\mathbf{Y} $ into $S$ subsets, $\{\mathbf{Y}^1, \cdots, \mathbf{Y}^S  \}$, where $S = \lceil N/1000 \rceil$. 
\For{$s = 1$ to $S$}
\State Build a CDMMA, $\mathcal{M}^s$, by applying Algorithm~\ref{algorithm_DSFMA} on $\mathbf{Y}^s$ taking $n$ as the subspace dimension; maximum number of auxiliary points as equal to $r_{max} \times \#\mathbf{Y}^s$ (where $\#\mathbf{Y}^s$ is the number of data points in $\mathbf{Y}^s$); and $L$ as the number of layers.   
\EndFor
\State \Return the parameters set $\mathcal{P} = \{\mathcal{M}^s\}_{s=1}^S$.
\end{algorithmic}
\end{algorithm} 
\begin{algorithm}
\caption{Variational learning of the classifier~\cite{10.1007/978-3-030-87101-7_14,kumar2022differentially}}
\label{algorithm_classification}
\begin{algorithmic}[1]
\Require Labeled data set $\mathbf{Y} = \left \{ \mathbf{Y}_c\; | \; \mathbf{Y}_c =  \left \{ y^{i,c} \in \mathbb{R}^p \; | \; i \in \{1,\cdots,N_c \} \right \},\: c \in \{1,\cdots,C \} \right \}$; the subspace dimension $n \in \{1,\cdots,p\}$; ratio $r_{max} \in (0,1]$; the number of layers $L \in \mathbb{Z}_{+}$.
\For{$c = 1$ to $C$} 
\State Build a wide CDMMA, $\mathcal{P}_c = \{\mathcal{M}^s_c\}_{s=1}^{S_c}$, by applying Algorithm~\ref{algorithm_WDSFMA} on $\mathbf{Y}_c$ for the given $n$, $r_{max}$, and $L$.  
\EndFor
\State \Return the parameters set $ \{ \mathcal{P}_c  \}_{c=1}^C$. 
\end{algorithmic}
\end{algorithm} 
\begin{algorithm}
\caption{Differentially private approximation of data samples~\cite{kumar2022differentially}}
\label{algorithm_differential_private_approximation}
\begin{algorithmic}[1]
\Require  Data set $\mathbf{Y} =  \left \{ y^{i} \in \mathbb{R}^p \; | \; i \in \{1,\cdots,N \} \right \}$; differential privacy parameters: $d  \in \mathbb{R}_{+}$,  $\epsilon  \in \mathbb{R}_{+}$, $\delta \in (0,1)$.
\State A differentially private approximation of data samples is provided as
\begin{IEEEeqnarray}{rCl}
 y^{+i}_j & = & y^{i}_j + F_{\mathrm{v}_j^{i}}^{-1}(r^{i}_j;\epsilon,\delta,d),\; r^{i}_j \in (0,1) \\
 \label{eq_inverse_cdf}F_{\mathrm{v}_j^i}^{-1}(r^i_j;\epsilon,\delta,d) & = & \left \{\begin{array}{ll} \frac{d}{\epsilon} \log(\frac{2r^i_j}{1 - \delta}), & r^i_j <  \frac{1- \delta}{2} \\
0, &  r^i_j \in [ \frac{1- \delta}{2}, \frac{1+ \delta}{2}]  \\
 -\frac{d}{\epsilon} \log(\frac{2(1-r^i_j)}{1-\delta}), & r^i_j > \frac{1+\delta}{2}
\end{array} \right.,\; r^i_j \in (0,1).
\end{IEEEeqnarray}     
where $ y^{+i}_j$ is $j-$th element of  $ y^{+i} \in \mathbb{R}^p$.    
\State \Return $\mathbf{Y}^+ = \left \{ y^{+i} \in \mathbb{R}^p \; | \; i \in \{1,\cdots,N \} \right \}$. 
\end{algorithmic}
\end{algorithm} 
\begin{algorithm}
\caption{Variational learning of a differentially private classifier~\cite{kumar2022differentially}}
\label{algorithm_private_classification}
\begin{algorithmic}[1]
\Require Differentially private approximated dataset: $\mathbf{Y}^+ = \left \{ \mathbf{Y}_c^+\; | \;  c \in \{1,\cdots,C \} \right \}$; the subspace dimension $n \in \{1,\cdots,p\}$; ratio $r_{max} \in (0,1]$; the number of layers $L \in \mathbb{Z}_{+}$.  
\State Build a classifier, $ \{ \mathcal{P}_c^+  \}_{c=1}^C$, by applying Algorithm~\ref{algorithm_classification} on $\mathbf{Y}^+$ for the given $n$, $r_{max}$, and $L$.  
\State \Return $ \{ \mathcal{P}_c^+  \}_{c=1}^C$. 
\end{algorithmic}
\end{algorithm}

\subsection*{Author contributions}

This is an author contribution text. This is an author contribution text. This is an author contribution text. This is an author contribution text. This is an author contribution text. 

\subsection*{Financial disclosure}

None reported.

\subsection*{Conflict of interest}

The authors declare no potential conflict of interests.

\section*{Supporting information}

\appendix

\bibliography{bibliography}%

\clearpage

\end{document}